\documentclass[conference]{IEEEtran}
\usepackage{times}

\usepackage[numbers]{natbib}
\usepackage{amsmath}
\usepackage{amsfonts}
\usepackage{amssymb}
\usepackage{multicol}
\usepackage[bookmarks=true]{hyperref}
\usepackage{graphicx}   
\usepackage{caption}    
\usepackage{subcaption} 
\usepackage{algorithm}
\usepackage{algpseudocode}
\usepackage{booktabs}
\DeclareMathOperator*{\argmax}{arg\,max}
\DeclareMathOperator*{\argmin}{arg\,min}

\begin{document}

\title{Constrained Gaussian Process Motion Planning via Stein Variational Newton Inference}

% You will get a Paper-ID when submitting a pdf file to the conference system
% \author{Jiayun Li, Kay Pompetzki, An Thai Le, Haolei Tong, Jan Peters, Georgia Chalvatzaki}
\author{
  Jiayun Li \quad
  Kay Pompetzki \quad
  An Thai Le \quad
  Haolei Tong \quad
  Jan Peters \quad
  Georgia Chalvatzaki
}

% \author{\authorblockN{Michael Shell}
% \authorblockA{School of Electrical and\\Computer Engineering\\
% Georgia Institute of Technology\\
% Atlanta, Georgia 30332--0250\\
% Email: mshell@ece.gatech.edu}
% \and
% \authorblockN{Homer Simpson}
% \authorblockA{Twentieth Century Fox\\
% Email: homer@thesimpsons.com}
% \and
% \authorblockN{James Kirk\\ and Montgomery Scott}
% \authorblockA{Starfleet Academy\\
% San Francisco, California 96678-2391\\
% Telephone: (800) 555--1212\\

% avoiding spaces at the end of the author lines is not a problem with
% conference papers because we don't use \thanks or \IEEEmembership

%Homer Simpson\authorrefmark{2},
%Montgomery Scott\authorrefmark{3} and
%Eldon Tyrell\authorrefmark{4}}
%Telephone: (800) 555--1212, Fax: (888) 555--1212}
%\authorblockA{\authorrefmark{4}Tyrell Inc., 123 Replicant Street, Los Angeles, California 90210--4321}}

\maketitle

\vspace{-2ex}
\renewcommand{\thefootnote}{\fnsymbol{footnote}}
\footnotetext[1]{ \noindent  
Computer Science Department, Technische Universität Darmstadt \&  Hessian.AI, Darmstadt, Germany. Email: jiayun.li@tu-darmstadt.de}

\begin{abstract}
Gaussian Process Motion Planning (GPMP) is a widely used framework for generating smooth trajectories within a limited compute time--an essential requirement in many robotic applications. However, traditional GPMP approaches often struggle with enforcing hard nonlinear constraints and rely on Maximum a Posteriori (MAP) solutions that disregard the full Bayesian posterior. This limits planning diversity and ultimately hampers decision-making. Recent efforts to integrate Stein Variational Gradient Descent (SVGD) into motion planning have shown promise in handling complex constraints. Nonetheless, these methods still face persistent challenges, such as difficulties in strictly enforcing constraints and inefficiencies when the probabilistic inference problem is poorly conditioned.
To address these issues, we propose a novel constrained Stein Variational Gaussian Process Motion Planning (cSGPMP) framework, incorporating a GPMP prior specifically designed for trajectory optimization under hard constraints. Our approach improves the efficiency of particle-based inference while explicitly handling nonlinear constraints. This advancement significantly broadens the applicability of GPMP to motion planning scenarios demanding robust Bayesian inference, strict constraint adherence, and computational efficiency within a limited time.  We validate our method on standard benchmarks, achieving an average success rate of 98.57\% across 350 planning tasks, significantly outperforming competitive baselines. This demonstrates the ability of our method to discover and use diverse trajectory modes, enhancing flexibility and adaptability in complex environments, and delivering significant improvements over standard baselines without incurring major computational costs. %We validate our planner through both simulated and physical experiments.

\end{abstract}

\IEEEpeerreviewmaketitle

\section{Introduction}
%% motivate the Stein method and constraints
Motion planning is a fundamental component of robotic manipulation and mobile robot navigation. While both optimization-based \cite{schulman2014motion, DBLP:journals/corr/Toussaint14} and sampling-based motion planners \cite{gammell2014informed, 5980479} have achieved significant successes, e.g., in producing smooth trajectories and handling non-holonomic constraints respectively, each approach has inherent limitations. Optimization-based planners excel at generating locally optimal trajectories and explicitly handling constraints but are heavily dependent on initialization, often converging to local optima \cite{zucker2013chomp}. In contrast, sampling-based planners offer probabilistic completeness and can discover globally optimal modes, but they struggle to efficiently produce high-quality trajectories that adhere to optimal objectives and constraints \cite{gammell2014informed, karaman2011sampling}. These challenges motivate exploring methods that combine the strengths of both approaches to develop a probabilistically complete framework while also adhering to optimal objectives and strict constraints \cite{9560731}.

\begin{figure}[t!]
    \centering
    % \vspace{-5pt}
    \includegraphics[width=1.0\linewidth]{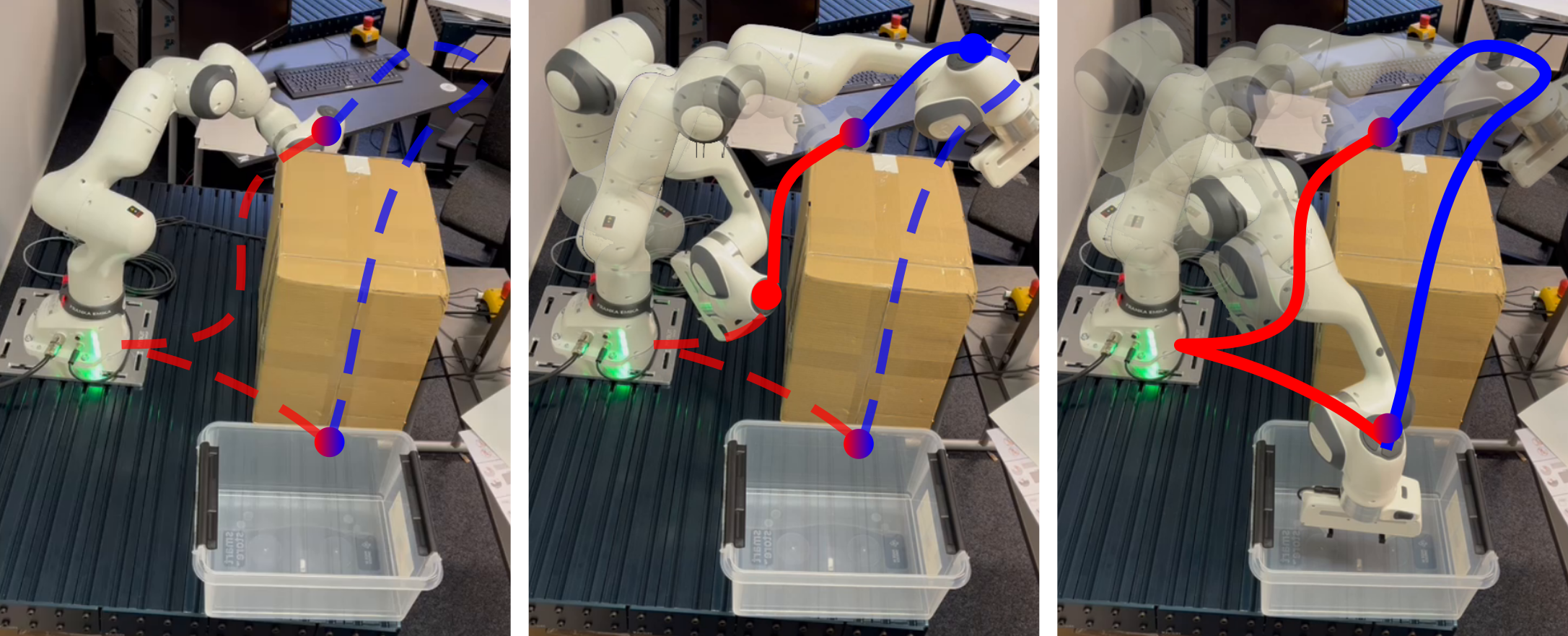}
    \caption{The figure illustrates a real-world demonstration of the robot executing two distinct trajectory modes identified by our planner to achieve the same task. The solid line represents the motion of the Panda hand.}
    \label{fig:real_exp}
    \vspace{-12pt}
\end{figure}

%% motivate the regularization prior & SVN
Unfortunately, tackling high-dimensional sampling problems with nonlinear constraints requires more sophisticated solutions.
Solving nonlinear programming problems through sampling has a long history but remains limited by their inherently sequential nature, which often fails to explore complex, high-dimensional solution spaces \cite{girolami2011riemann}. This approach typically struggles with multimodal distributions and disconnected feasible regions, limiting its ability to achieve efficient exploration and effectively handle nonlinear constraints \cite{berenson2009manipulation, kalakrishnan2011stomp, le2024accelerating, le2024global}. Stein Variational Gradient Descent (SVGD) has emerged as a promising tool for trajectory optimization (TO) within probabilistic inference frameworks \cite{lambert2020stein, lee2024stein}. However, its application in motion planning is hindered due to inefficient score-based inference in ill-conditioned problems. Additionally, there are no effective methods to handle strict nonlinear constraints efficiently. As a result, runtimes become prohibitively long, making it unsuitable for scenarios with tight compute time budgets. Furthermore, the diversity-driven mechanism of SVGD is highly sensitive to the chosen trajectory distance metric; conventional point-to-point distance metrics often reduce the algorithm's ability to effectively explore the state space. While annealing schemes frequently encourage particles to spread homogeneously across the state-space initially, this approach risks driving particles too far from regions of interest, ultimately undermining the planning performance. Therefore, prior work \cite{lambert2021entropy} suggests that a well-defined probabilistic trajectory prior is crucial for providing a regularized exploration space, effectively guiding the behavior of particles, and enhancing their efficiency in motion planning tasks. Note that random particle initialization followed by posterior sampling without regularization is prone to fail~\cite{toussaint2024nlp}.

Gaussian Process Motion Planning (GPMP) is a framework that reinterprets trajectory optimization as a probabilistic inference problem on a factor graph derived from a linear Stochastic Differential Equation (SDE). This approach exploits the sparse structure of the inference problem on the factor graph, derived from the Markov assumption, to efficiently compute the Maximum a-Posteriori (MAP) solution \cite{mukadam2018continuous}. While GPMP offers significant advantages in computational efficiency and robustness, its reliance on a linear prior poses challenges when addressing nonlinear programming tasks. Specifically, when confronted with nonlinear constraints, the inference process becomes incompatible because the hypothesis space is constrained by the linear prior. Moreover, the distance metric in the Reproducing Kernel Hilbert Space (RKHS), induced by the linear SDE kernel, is limited in the scope of neighboring elements due to its diagonally banded structure. Consequently, integrating a nonlinear prior is essential for enabling GPMP to effectively handle scenarios with nonlinear constraints.

To address the aforementioned limitations, we propose Constrained Stein Variational GPMP (cSGPMP). This approach solves the GPMP problem with constraints using a novel method called Constrained Stein Variational Newton (cSVN). In the GPMP framework, we introduce an analytically integrable nonlinear GP prior to the highest-level trajectory of interest, such as acceleration. This enables the computation of the lower-level trajectory prior and their joint distribution analytically. The proposed prior not only provides a hypothesis space encompassing trajectories compliant with nonlinear constrants but also serves as a well-defined distance metric for Stein variational inference. Subsequently, we formulate the nonlinear constrained variational inference problem as a Sequential Quadratic Programming (SQP) problem. Each update involves solving an equality-constrained Stein Variational Newton (SVN) step, ensuring accurate and efficient inference under constraints. We validate the proposed planner through simulation benchmarks for both mobile navigation and manipulation tasks, as well as real-world experiments with a 7-DoF manipulator robot operating in constrained environments. Our experiments demonstrate the efficiency and capability of cSGPMP to generate high-quality trajectories compliant with constraints. Our planner significantly improves planning outcomes by achieving a higher success rate and finding better objective function solutions compared to other baselines, all while adhering more effectively to constraints.

% First, we introduce a constrained Stein Variational Newton (cSVN) method for efficient sampling from constrained state spaces, which significantly enhances inference speed and robustness compared to existing probabilistic inference-based planners, while ensuring strict adherence to hard constraints. Second, we extend GPMP with a generalized nonlinear prior, enabling compatibility with nonlinear constraints and improving coherence in complex scenarios. The resulting RKHS serves as a well-defined distance metric and provides an effective kernel for cSVN. We validate the proposed planner through simulation benchmarks both for particle-based planning and manipulation tasks, and in real-world experiments with a 7DoF manipulator robot in constrained environments. Our experiments demonstrate the efficiency and capability of cSGPMP to generate high-quality trajectories compliant with constraints. 

In summary, our contributions are twofold:
\begin{itemize}
    \item[(i)] We propose a Constrained Stein Variational Newton method for efficient sampling in constrained state spaces. This method significantly enhances inference speed and robustness compared to existing gradient-based and inference-based planners, while guaranteeing strict adherence to hard constraints.
    \item[(ii)] We extend GPMP by incorporating a generalized nonlinear prior, enabling compatibility with nonlinear constraints and enhancing coherence in complex scenarios. Its RKHS serves as a well-defined distance metric for trajectories, providing a foundation for the cSVN method.
\end{itemize}
\section{Related Works}

\subsection{Particle-based methods for TO}
Recent efforts in particle-based TO have focused on leveraging mutual information among particles or learning vector fields from the posterior. These approaches show the potential to achieve better local optima with a batch of particles. Lambert et al. \cite{lambert2020stein} utilized SVGD to approximate posterior distributions for trajectory optimization, using initialization as sampling from trajectory prior. This method enables multimodal trajectory sampling for model predictive control tasks but treats SVGD solely as a diversity-driven mechanism rather than a full Bayesian inference tool. To address this limitation, Lambert et al. \cite{lambert2021entropy} introduced a vanilla GPMP prior as a regularizer, improving inference but leaving constraints handling and convergence speed unaddressed.

Recent advances in improving convergence speed include the use of Stein Variational Newton (SVN) \cite{aoyama2024second} and optimal transport methods \cite{le2024accelerating} to accelerate inference, but both approaches struggle with constraints handling. Power et al. \cite{power2024constrained} proposed a constrained SVGD framework that employs an orthogonal kernel trick to project the kernel gradient and score onto the tangent space of the constraints, ensuring safety during optimization. While this method enforces strict satisfaction of constraints by projecting SVGD to the tangent space of constraints, it struggles with particle movement along the constraints due to the lack of curvature information about the log-likelihood function (cf.~Sec.~\ref{subsec:constrained_svgd_svn}).

\subsection{Sampling on Constraint Manifolds}

Motion planning on constraint manifolds remains a pivotal research area with significant implications for robotics. Suh et al. \cite{suh2011tangent} introduced a randomized approach that relies on frequent projection operations to maintain tree structures in high-dimensional spaces, incurring high computational costs. Tangent Bundle RRT \cite{kim2016tangent} improves efficiency by constructing random trees on tangent bundles, but struggles with complex or highly nonlinear constraints, often facing convergence issues. Toussaint et al. \cite{toussaint2024nlp} proposed a two-stage method that satisfies constraints using the Gauss-Newton method, followed by hit-and-run interior sampling. This approach avoids the complexities of the Lagrangian multiplier framework while achieving diverse, constraint-adhering sampling. The study also provides a comparison of manifold RRT and Riemannian Langevin dynamics for sequential sampling. However, despite the effectiveness of those methods, they rely on stochastic sampling, requiring sufficient Markov Chain mixing. Additionally, the sampled points lack mutual information sharing, which hinders fast convergence. Recent works~\cite{ortiz2022structured,diamond2023geometric,acar2021approximating}, employ structured deep generative models for constraint manifold sampling in manipulation tasks. While effective within learned distributions, these methods struggle to generalize to unseen constraints and demand extensive training data, incurring high computational costs.
 
\section{Preliminaries}
% Here, we introduce the fundamentals of SVGD and SVN, including the formulation of the GPMP inference problem.

\subsection{Stein Variational Gradient Descent \& Stein Variational Newton}
Stein's method provides a robust probabilistic distance measure suitable for variational inference when applied to two different probability distributions $q$ and $p$. To make this concept numerically realizable, the kernelized Stein discrepancy has been proposed
\begin{equation}
\label{eq:ksd}
    \mathbb{S}(q,p) = \max_{\boldsymbol{\phi}\in \mathcal{H}^d} \mathbb{E}_{x\sim q} [\text{Tr}[\mathcal{A}_p \boldsymbol{\phi}(x) ]], \quad s.t. \,||\boldsymbol{\phi}||_{\mathcal{H}} \leq 1.
\end{equation}
Here, $\boldsymbol{\phi} \in \mathcal{H}^d$ is an arbitrary smooth vector function in a RKHS, and \(\mathcal{A}_p\) is Stein's operator. When sampling from $p(x)$, Stein's operator transforms \(\boldsymbol{\phi}\) into a zero-mean function defined as \(\boldsymbol{\phi}(x) \nabla_x \log p(x)^T + \nabla_x \boldsymbol{\phi}(x)\). This makes it a valid distance metric, as it equals 0 when the two distributions \(q\) and \(p\) are identical. We use \(\nabla_x\) to denote the gradient w.r.t. the variable \(x\), and \(\text{Tr}\) to denote the trace, while $\mathcal{H}^d$ is a vector-valued RKHS induced by a positive-definite kernel function $k$. It satisfies the reproducing property: \(\langle k(x, \cdot), f \rangle_{\mathcal{H}} = f(x)\), \(\langle k(x, \cdot), k(y, \cdot) \rangle_{\mathcal{H}} = k(x, y)\) and \(\nabla_x f(x) = \langle f, \nabla_x k(x,\cdot)\rangle_{\mathcal{H}}\), where $\langle\cdot,\cdot \rangle_{\mathcal{H}}$ is the inner-product in RKHS.

It can be shown that for small perturbations, where there exists a transport map \(\mathbf{T}\) such that \(\mathbf{T}(x) = x + \epsilon \boldsymbol{\phi}(x)\) as \(\epsilon \to 0\), the function \(\boldsymbol{\phi}\) in Eq. \ref{eq:ksd} corresponds to the steepest descent direction for the Kullback-Leibler (KL) divergence. Specifically, it maximizes \(-\nabla_\epsilon \text{KL}[q(\mathbf{T}[x]) || p(x)]\). This steepest descent direction is referred to as SVGD~\cite{liu2016stein}, which has the following functional gradient form
\begin{equation}
    \boldsymbol{\phi} = \mathbb{E}_{x\sim q} [\nabla_x \log p(x) k(x,\cdot) + \nabla_x k(x,\cdot)].
\end{equation}

Furthermore, Stein Variational Newton (SVN) represents a second-order approximation of the variation in the KL divergence. It is expressed as a functional positive-definite operator analogous to the Hessian matrix in Euclidean space, with the form:
\begin{equation}
    \label{eq:svn}
    \mathbf{H} = \mathbb{E}_{x\sim q} [\nabla^2_x \log p(x) k(x,\cdot)k(x,\cdot) + \nabla_xk(x,\cdot) \nabla_x k(x,\cdot)^T].
\end{equation}
By solving the linear system constructed from the numerical estimation of \(\mathbf{H}(y, z)\), representing the \((y, z)\)-block of a block square matrix for all \(y, z \in Q\), and \(\boldsymbol{\phi}(y)\), representing the \(y\)-th entry of a tall vector for all \(y \in Q\), one can form a linear system using a set of particles \(Q\). The solution to this system corresponds to the Stein Variational Newton step \cite{detommaso2018stein}.

\subsection{Gaussian Process Motion Planning}
GPMP is a continuous-time local planner that can generate smooth trajectories through a GP prior regularization while adding obstacle avoidance as a likelihood function. The GP prior used in GPMP is the linear SDE which has the following form
\begin{equation}
    \dot{\xi}(t) = A(t) \xi(t) + u(t) + F(t) w(t)
\end{equation}
where the $A(t), F(t)$ are the system matrix and noise input matrix, $u(t)$ is the control input, and $w(t)$ is a white noise process. This prior will induce a GP prior $p(\xi) \propto \exp(-\frac{1}{2} ||\xi - \mu||^{2}_{\mathcal{K}})$ along a time horizon, e.g., $t \in [0,1]$, and with $\mu$ being $\mathbb{E}[\xi]$ and $\mathcal{K}$ the $\text{Cov}[\xi]$. If we have a set of support points to realize the GP trajectory $p(\xi)$, the covariance of the GP trajectory $\text{Cov}[\xi]$ has a diagonal-banded structure, therefore, the MAP optimization (\ref{eq:map_problem}) can be accelerated by utilizing sparsity,
\begin{align}
\label{eq:map_problem}
    \xi^{*} &= \argmax_{\xi} \, p(\xi) l(\xi; \theta),
\end{align}
where \( l(\xi; \theta) \) represents the likelihood function parameterized by \(\theta\). It typically takes the form of a Boltzmann distribution, given by \( l(\xi; \theta) \propto \exp(-L(\xi; \theta)) \), where \(L\) represents the loss function to be minimized. After applying a logarithm transformation, the MAP problem is equivalent to minimizing the negative log posterior as
\begin{equation}
\label{eq:log_map}
    \xi^{*} = \argmin_{\xi} \frac{1}{2}||\xi - \mu||_{\mathcal{K}}^2 + L(\xi;\theta).
\end{equation}
For more details, we refer to \cite{mukadam2018continuous} and Chapter 3.4~\cite{barfoot2024state}.
\section{Constrained Stein Variational Newton GPMP}
In this section, we introduce the key components of cSGPMP, including the formulation of the nonlinear GPMP prior and the equality-constrained SVGD / SVN step in trajectory space.

\subsection{Problem Definition}
Instead of solving the GPMP problem through MAP (Eq.~\ref{eq:map_problem}), we formulate it as a variational inference problem that minimizes the KL divergence between the posterior and our variational distribution 
\begin{align}
    \min_{q(\xi)}\quad &\text{KL} \Big[q(\xi) \big|\big| p(\xi) l(\xi,\theta)\Big],\\
    \nonumber\text{s.t.}& \quad \mathbb{E}_{\xi \sim q} [h(\xi)] = 0,
\end{align}
where the variational distribution $q$ is represented as a set of particles.
We formulate the variational inference problem with constraints $h(\xi)$ as a Sequential Quadratic Programming (SQP) optimization task. Each iteration of the constrained Stein Variational Gradient Descent (SVGD) or Stein Variational Newton (SVN) method is formulated as an equality-constrained Linear Programming (LP) or Quadratic Programming (QP) problem. In this work, we do not explore inference tasks with inequality constraints, as these significantly increase the complexity of satisfying the Karush-Kuhn-Tucker (KKT) conditions, which grow combinatorially. Moreover, in most cases, inequality constraints can be transformed into equality constraints by introducing slack variables. For completeness, we provide details on translating inequality constraints into equality constraints and the corresponding modified KKT system in the Appendix.

\subsection{A general nonlinear GPMP prior}
\label{subsec:gpmp_prior}
\begin{figure}[t]
  \centering
  \vspace{-3pt}
  \begin{subfigure}[t]{0.235\textwidth}
    \centering
    \includegraphics[width=\textwidth]{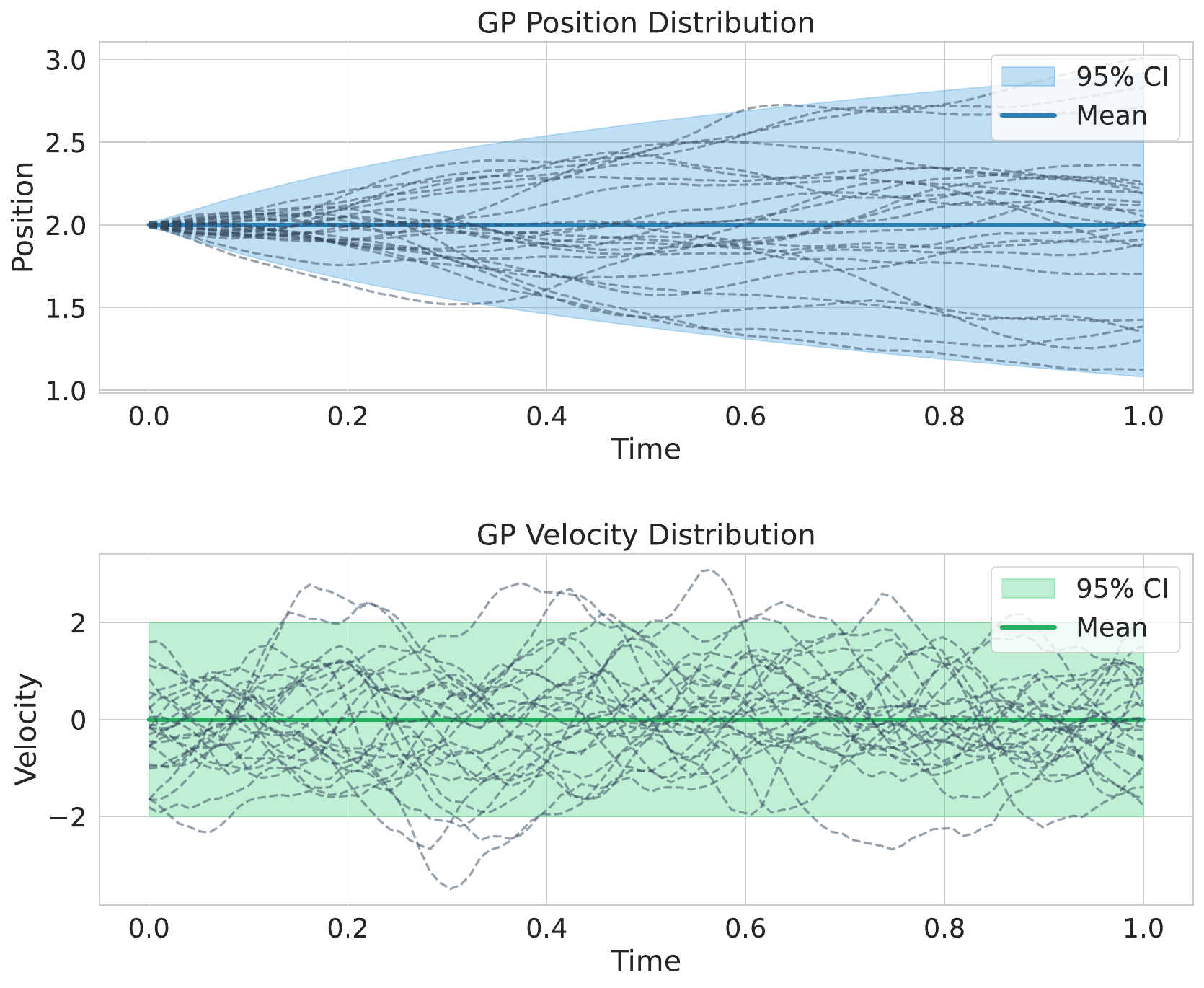}
    \captionsetup{labelformat=empty} % Remove subfigure labels
    \caption{The GP prior.}
    \label{fig:subfig1}
  \end{subfigure}
  \hspace{0.001\textwidth} % Reduce horizontal spacing
  \begin{subfigure}[t]{0.235\textwidth}
    \centering
    \includegraphics[width=\textwidth]{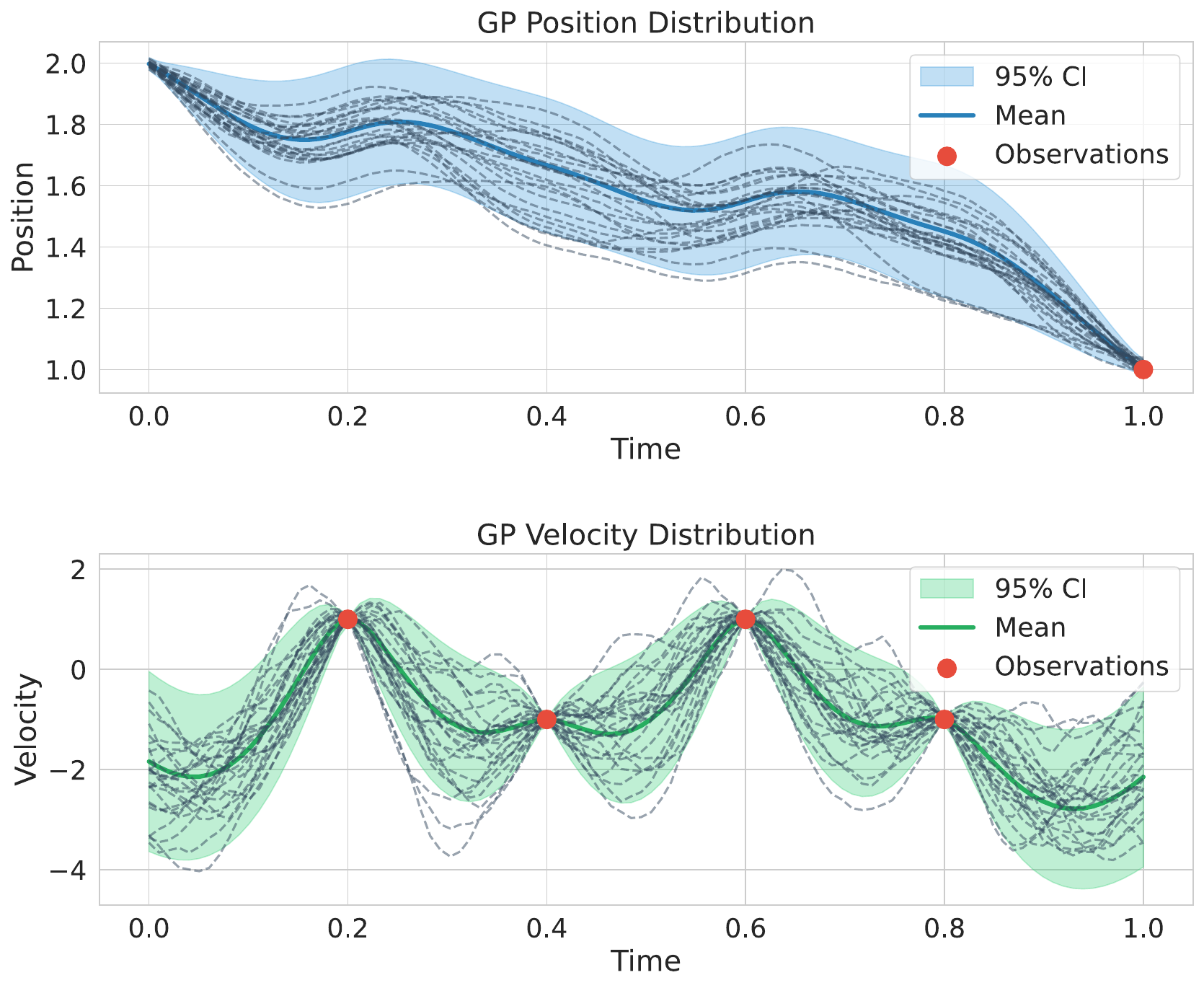}
    \captionsetup{labelformat=empty} % Remove subfigure labels
    \caption{GP Posterior.}
    \label{fig:subfig2}
  \end{subfigure}
  \caption{We set \(x_0\) to be \(p(x_0) 
  \sim \mathcal{N}(2, 10^{-4})\). The kernel for velocity is Matérn 3/2. The GP prior represents the distribution without any observations, while the GP posterior is computed using asynchronous observations of position and velocity. The velocity noise \(\sigma_n^2\) is also set to \(10^{-4}\). This process is regarded as a nonlinear continuous-time diffusion starting from \( x(0) = 2 \). With observations, the diffusion uncertainty is reduced.}
  \label{fig:hsgp_demo}
  \vspace{-10pt}
\end{figure}

To ensure that the variational inference has a robust probabilistic prior, we introduce a nonlinear GP trajectory prior to enhance the robustness of the inference process. Specifically, we propose a GP prior based on spectral decomposition within the Hilbert-space Gaussian Process (HSGP) regression framework \cite{solin2020hilbert}. This approach is applicable to any stationary kernel, which is approximated using features $\varphi$ similar to the Fourier ones and its eigenvalues $S(\sqrt{\lambda})$. The $S$ is the spectral density of the kernel function that we approximate. In this work, we demonstrate only the derivation of the highest trajectory level at velocity. By directly integrating a Gaussian Process velocity prior into the position, we obtain a joint GP distribution, enabling joint inference. This approach is well-suited for generating physically consistent trajectories that adhere to nonlinear constraints.
Therefore, we introduce the velocity prior as a scalar GP, denoted as \( v(t) \), with time input \( t \), defined as:
\begin{align}
    v(t) &\sim \mathcal{N}\,
    (m(t), k(t, t')+ \sigma_n^2 \textbf{1}(t-t')), \\
    m(t) &= (x_{T} - x_{0}) / T, \\
    k(t,t') &= \sum_j S(\sqrt{\lambda_j})\varphi_j(t) \varphi_j(t').
\end{align}
The Dirac delta function, denoted as \( \textbf{1}(\cdot) \), is defined such that \( \textbf{1}(x) = 1 \) when \( x = 0 \). \( T \) represents the time point of trajectory state $x$ at the end of the trajectory. Consequently, \( m(t) \) serves as a constant velocity prior, although a zero-mean velocity prior can also be used. The introduction of HSGP is necessary as we aim to avoid having the GP prior act on the entire state-space, hence reducing the hypothesis space. It is also essential to separate the input points to enable explicit integration within the feature space. Thus, the distribution of the position trajectory is
\begin{align}
    x(t) &= x_0 + \int_0^t v(\tau) d\tau \\
    \mathbb{E}[x(t)] &= \mathbb{E}[x_{0}] + \int_{0}^{t} m(\tau) d\tau \\
    \text{Cov} (x(t), x(s)) &= var(x_0) + \int_0^t \int_0^s k(\tau, \tau') d\tau d\tau' \\&+ \min(t,s) \sigma^2_n \notag
\end{align}
The correlation bewteen $x$ and $v$ is
\begin{equation}
    \text{Cov}(x(t), v(s)) = \int_0^t k(\tau, s) d\tau
\end{equation}
We give details of analytically integration of the kernel function in Appendix \ref{subsec:hsgp_integration}. The joint distribution can now be derived, and to make predictions we can use the marginalized GP distribution (more details in~\cite{rasmussen2003gaussian}),
\begin{equation}
        p(\xi) = \mathcal{N}\Biggl(\begin{bmatrix}
        x \\v
    \end{bmatrix},
    \begin{bmatrix}
        \text{Cov}(x,x) & \text{Cov}(x,v) \\
        \text{Cov}(v,x) & \text{Cov}(v,v) \\
    \end{bmatrix}\Biggl)
\end{equation}
A simple 1-D example for the joint prior distribution is illustrated in Fig. \ref{fig:hsgp_demo}. As shown, samples drawn from this prior naturally satisfy the differential relationships between position and velocity. This behavior can be attributed to the integration being performed in the feature or frequency domain. Furthermore, the uncertainty is correctly propagated, providing a robust trajectory prior to subsequent variational inference when prior knowledge, such as via points, is available. We adopt the approach of treating each joint independently. Achieving a vector-valued version that incorporates the correlations among all joints can be easily accomplished by following the method outlined in \cite{solin2020hilbert}. %We apply this prior individually to each joint; obtaining a vector-valued version can be straightforwardly achieved by following the approach in \cite{solin2020hilbert}.

\subsection{Equality-constrained SVGD and SVN}
\label{subsec:constrained_svgd_svn}
In the context of equality-constrained SVGD, the optimization problem can be formulated as aligning the steepest descent direction $\boldsymbol{\phi}$ (i.e., SVGD) with the target direction, subject to a linearized equality constraint with input $y$. This alignment is represented through the inner product $\langle\boldsymbol{\phi}, \delta \rangle_{\mathcal{H}}$ in RKHS, ensuring compliance with the constraint while optimizing the objective to get the optimal update direction $\delta(y)$. With linearized equality constraints, the problem is stated as
\begin{align}
    \min_{\delta} \quad \mathcal{L}_{1}(\delta) &= - \mathbb{E}_{x \sim q} [\text{Tr}[\nabla_x \log p(x) \delta(x)^T + \nabla_x \delta(x)]], \nonumber \\
    \text{s.t.}& \quad \mathbb{E}_{y} [\nabla h(y)^T \delta(y) + h(y)] = 0.
\end{align}
The Lagrangian function of this problem is 
\begin{align}
        \min_{\delta} \quad \mathcal{L} &= \mathcal{L}_{1}(\delta) +  \mathcal{L}_{cons}(\delta), \nonumber\\
       \text{where } \mathcal{L}_{cons} &= \boldsymbol{\lambda}^T \mathbb{E}_y [\nabla h(y)^T \delta(y) + h(y)].
\end{align}
 Using the reproducing properties, we can derive the functional gradient of this unconstrained problem as 
\begin{align}
    \nabla_{\delta} \mathcal{L} = - \boldsymbol{\phi} + \mu_p \nabla h \, \boldsymbol{\lambda}
\end{align}
where
\begin{align}
\label{eq:svgd}
    \boldsymbol{\phi} &= \mathbb{E}_x [\nabla_x \log p(x) k(x,\cdot) + \nabla_x k(x,\cdot) ]\\
    \label{eq:constraints_embedding}
    \mu_{p} \nabla h &= \mathbb{E}_y [\nabla h(y) k(y, \cdot)]
\end{align}
We can see the constraints now are represented as a Jacobian weighted \textit{kernel mean embedding} (denoted as $\mu_p$) in the RKHS. Without accounting for the input noise in \( y \), we can eliminate the expectation w.r.t. \( y \). Consequently, the KKT system is:
\begin{align}
\label{eq:complement_condition}
    \mathbb{E}_x[\nabla_x \log p(x) k(x,y) + \nabla_x k(x,y)]  &= \nabla h(y) \boldsymbol{\lambda} k(y, y) \\
\label{eq:primary_condition}
    \nabla h(y)^T \delta(y) + h(y) &= 0,
\end{align}
meaning that for any possible random variables $x$ and input $y$, this relationship should hold. By constructing the general solution of the KKT system, we can obtain the desired result % TODO: need a proof
\begin{align}
    \delta(y) &= \mathbb{E}_x [P(y) \log p(x) k(x,y) + P(y) \nabla k(x,y)] \nonumber\\
    & ~~~- (\nabla h(y)^{T})^{\dagger} h(y),\\
    P(y) &= I - \nabla h(y)[\nabla h(y)^T\nabla h(y)]^{-1} \nabla h(y)^T,
\end{align}
where $P(y)$ is the null space projection matrix of the constraints w.r.t. the input $y$, and $(\nabla_x h(y)^T)^{\dagger}$ is the pseudo-inverse to find a special solution for Eq. \ref{eq:primary_condition}. We assume $\nabla h(y)^T$ is full rank. This result of constrained SVGD (cSVGD) implies that the support particles can only impact the input particles in the direction of the tangent space of the constraints while gradually approaching the equality constraints. This result is similar to the orthogonal kernel approach introduced in \cite{zhang2022sampling} used in \cite{power2024constrained} for trajectory optimization. 

\begin{figure}[t]
    \centering
    \vspace{-3pt}
    \begin{subfigure}{0.49\linewidth}
        \centering
        \includegraphics[width=\linewidth]{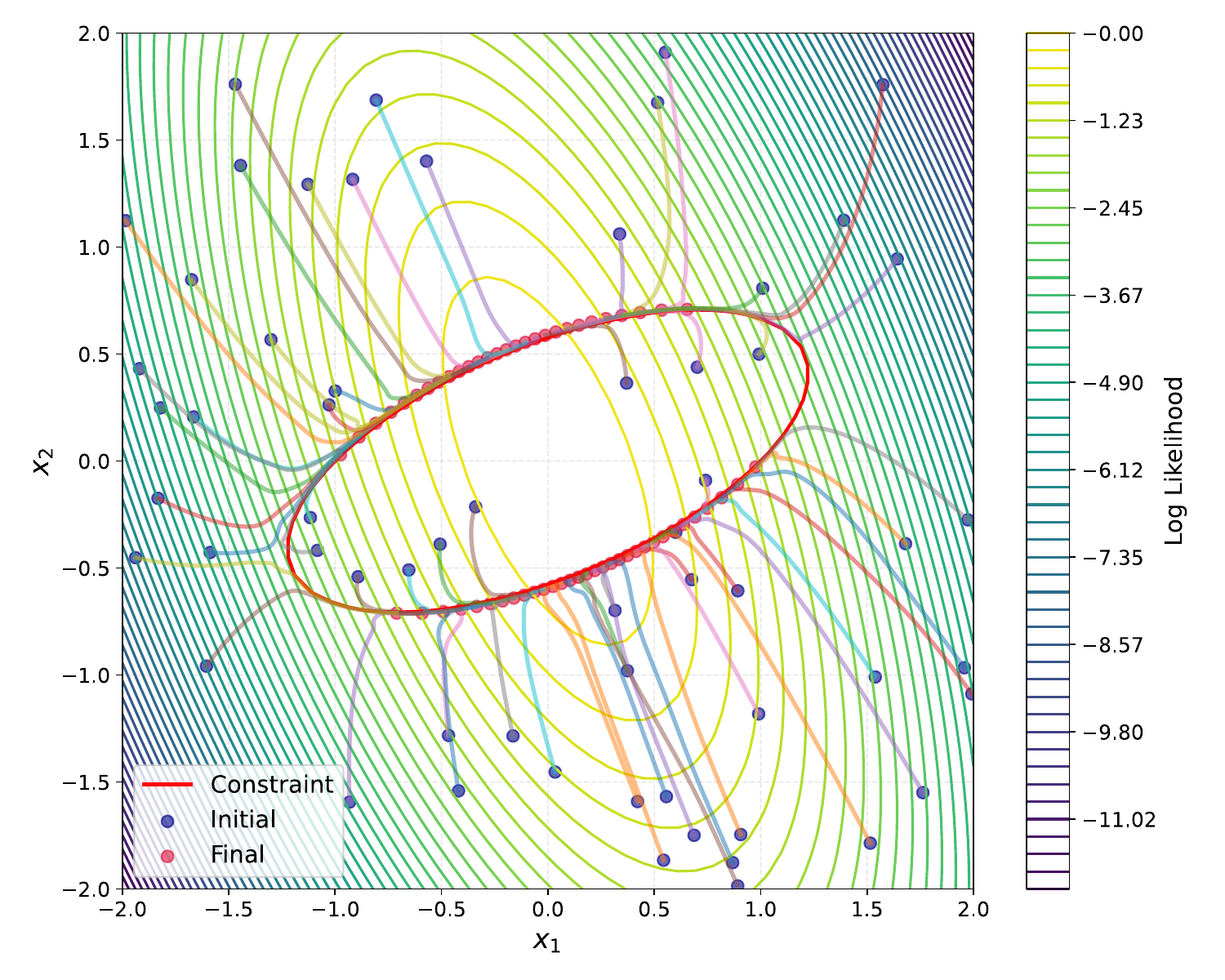}
        \caption{Constrained SVGD}
        \label{fig:csvn_gradient_flow}
    \end{subfigure}
    \begin{subfigure}{0.49\linewidth}
        \centering
        \includegraphics[width=\linewidth]{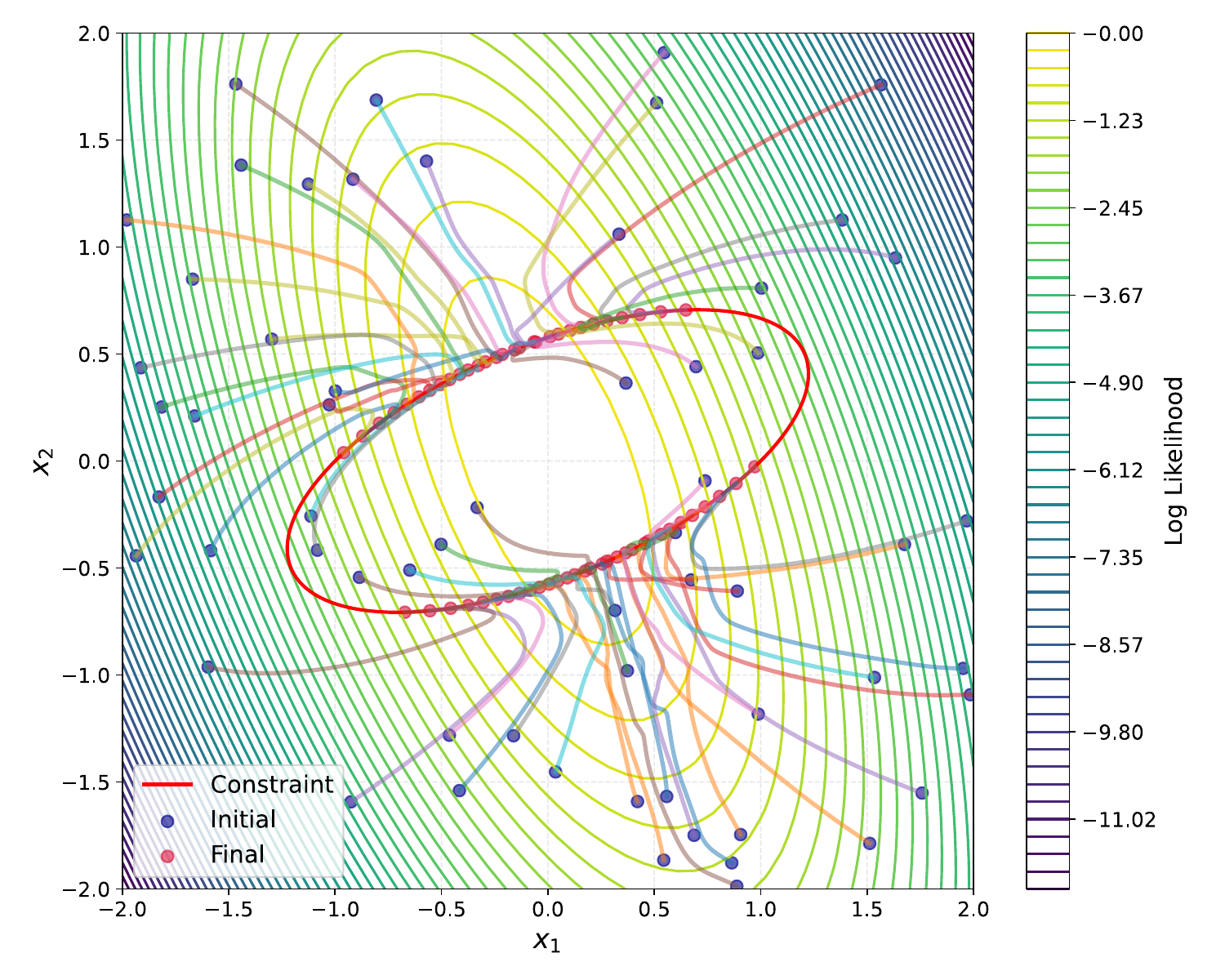}
        \caption{Constrained SVN}
        \label{fig:csvgd_gradient_flow}
    \end{subfigure}
    \caption{Particle trajectories for cSVGD and cSVN. The red ellipses represent an equality constraint, while the underlying distribution is a correlated Gaussian, shown in the background. The particles are initialized with the same random seed, and a small update rate of \(3e^{-5}\) is used to smooth the trajectories.}
    \label{fig:gradient_flow_comparison}
    \vspace{-5pt}
\end{figure}

To find the constrained solution of the SVN, we formulate the optimization problem as the second-order variation of the KL divergence \cite{detommaso2018stein}
\begin{align}
    \min_{\delta} &\quad \mathcal{L}_{1}(\delta) + \mathcal{L}_{2}(\delta) \nonumber\\
    \text{s.t.} \quad \mathbb{E}_{y}& [\nabla h(y)^T \delta(y) + h(y)] = 0
\end{align}
where $\mathcal{L}_{2}$ is
\begin{align}
    \mathcal{L}_{2} &= \frac{1}{2} \mathbb{E}_{x} [\delta(x)^T \nabla^2_x \log p(x) \delta(x) + \text{Tr}[\nabla_x\delta(x) \nabla_x\delta(x)]]
\end{align}
We can write down the Lagrangian function, using the reproducing properties as with the cSVGD we get  
\begin{equation}
\label{eq:complement_conditon_svn}
    \nabla_{\delta} \mathcal{L} = - \boldsymbol{\phi} + \langle \mathbf{H}, \delta \rangle_{\mathcal{H}} +  \mu_p \nabla h \, \boldsymbol{\lambda},
\end{equation}
where 
\begin{align}
\label{eq:stein_hessian}
    &\mathbf{H} = \mathbb{E}_x[-k(x,\cdot) \nabla^2_x \log p(x) k(x,\cdot) + \nabla_x k(x,\cdot)\nabla_x k(x,\cdot)^T]
\end{align}
We can also use reproducing property to write the equality constraints as 
\begin{equation}
\label{eq:primary_conditon_svn}
    \mathbb{E}_{y} [\langle \nabla h(y)^T k(y,\cdot), \delta\rangle_{\mathcal{H}}  + h(y)] = 0
\end{equation}
Because the complementary condition (Eq. \ref{eq:complement_conditon_svn}) and primary condition (Eq. \ref{eq:primary_conditon_svn}) are disentangled, assuming we have an input particle $y$, the KKT system is written as
\begin{align}
\label{eq:KKT_system}
    \begin{bmatrix}
    \mathbf{H} & \mu_p \nabla h \\
    \mu_p \nabla h^T & 0
\end{bmatrix}
\begin{bmatrix}
    \delta \\
    \boldsymbol{\lambda}
\end{bmatrix} + 
\begin{bmatrix}
    -\boldsymbol{\phi} \\
    \mathbb{E}_y[h(y)]
\end{bmatrix} =
0,
\end{align}
where the matrix multiplication is defined using the RKHS inner product as required. This linear system can be solved efficiently using Schur complement and potentially with the SVN block-diagonal approximation, thus can scale well w.r.t. the number of particles. This can be regarded as the Hessian-weighted projection, as in common equality-constrained QP problem. Fig. \ref{fig:gradient_flow_comparison} illustrates the trajectory for a simple example in which the optimization leverages local curvature information to accelerate convergence. This approach efficiently directs particles to the high-probability region while smoothly traversing the constraint manifold, avoiding fluctuations along the constraints seen in cSVGD.

We summarize the cSVN step in Algorithm \ref{alg:cSVN}. To achieve a numerically efficient block-diagonal cSVN step, we make the following three assumptions: (1) the input state is noise-free, (2) the Hessian matrix is block-diagonal approximated, and (3) the constraints evaluated at one particle can only affect the current particle. The derivation of the numerical solution for the KKT system, as discussed above, is provided in the Appendix \ref{subsec:numerical_cSGMP}. The cSVN method is designed to solve the GPMP variational inference problem, and its performance is demonstrated in Section \ref{sec:results}.

\begin{algorithm}[b]
\caption{Block-Diagonal cSVN Step}\label{alg:cSVN}
\begin{algorithmic}[1]
\Require A set of $N$ particles $Q$, a log-likelihood $\log p(x)$, a positive definite kernel $k$, an input $y$
\ForAll{$x_i \in Q$}
\State Get score function $\nabla \log p(x_i)$
\State Get (approximated) Hessian $\nabla^2\log p(x_i)$
\EndFor
\State Get constraint value $h(y)$ and gradient $\nabla h(y)$
\State Compute empirical SVGD with (\ref{eq:svgd}) \par
\noindent $\boldsymbol{\phi}(y) = \frac{1}{N} \sum^N_{i=1} \nabla_x \log p(x_i) k(x_i,y) + \nabla_x k(x_i, y) $
\State Compute empirical block-diagonal SVN with (\ref{eq:stein_hessian}) \par
\noindent $\mathbf{H} = \frac{1}{N}\sum^N_{i=1}[- \nabla^2 \log p(x_i) k(x_i,y)^2 + $\par
\hspace{4cm}$\nabla_x k(y,x_i) \nabla_x k(y,x_i)^T]$
\State Replace \(\mu_p \nabla h\) with \(\nabla h(y)\) and \(\mathbb{E}_y [h(y)]\) with \(h(y)\), and solve the KKT system (\ref{eq:KKT_system}) using the Schur complement with standard matrix algebra to compute \(\delta(y)\).
\State \Return $\delta(y)$
\end{algorithmic}
\end{algorithm}
As we have not accounted for the stochasticity of the input \( y \), the term \( \mu_p \nabla h \) simplifies to \( \nabla h(y) \). However, for applications where input uncertainty must be considered, the algorithm can be adapted accordingly.  

In this work, we use the BFGS method \cite{nocedal1999numerical} to approximate the second derivative of the log-likelihood \( \nabla^2_x \log p(x) \) for cSGPMP. To measure the distance between trajectory functions, we employ a generalized Radial Basis Function (RBF) kernel with the RKHS inner product induced by the GPMP prior introduced in section \ref{subsec:gpmp_prior}, formulated as  
\begin{equation}
k(\xi_i, \xi_j) = \sum^{n_q}_{k=1}\exp\left(-  \frac{1}{\sigma^{k2}} (\xi_i^k - \xi_j^k)^T\mathcal{K}^k(\xi_i^k - \xi_j^k)\right)
\end{equation}
where we assume there are $n_q$ joints and \( \sigma^k \) represents the kernel length for joint $k$. Unlike standard heuristics for using a dynamic kernel length w.r.t. the median distance between particles, we found that in our framework, kernel length plays a less significant role compared to SVGD, provided that it has the correct scaling in specific problems. To encourage exploration, we employ a warm-up phase with several dozen annealing steps. This approach generates sufficiently good results without introducing additional complexity into our framework.

\section{Results}

In this section, we evaluate the efficacy of our proposed cSVGD and cSVN by comparing them with standard trajectory optimizers in constrained motion planning tasks, including unicycle navigation and real-world high-dimensional manipulation scenarios.

\label{sec:results}
\subsection{Unicycle Navigation Task}
\begin{figure}[ht]
    \centering
    \vspace{-5pt}
    \includegraphics[width=0.85\linewidth]{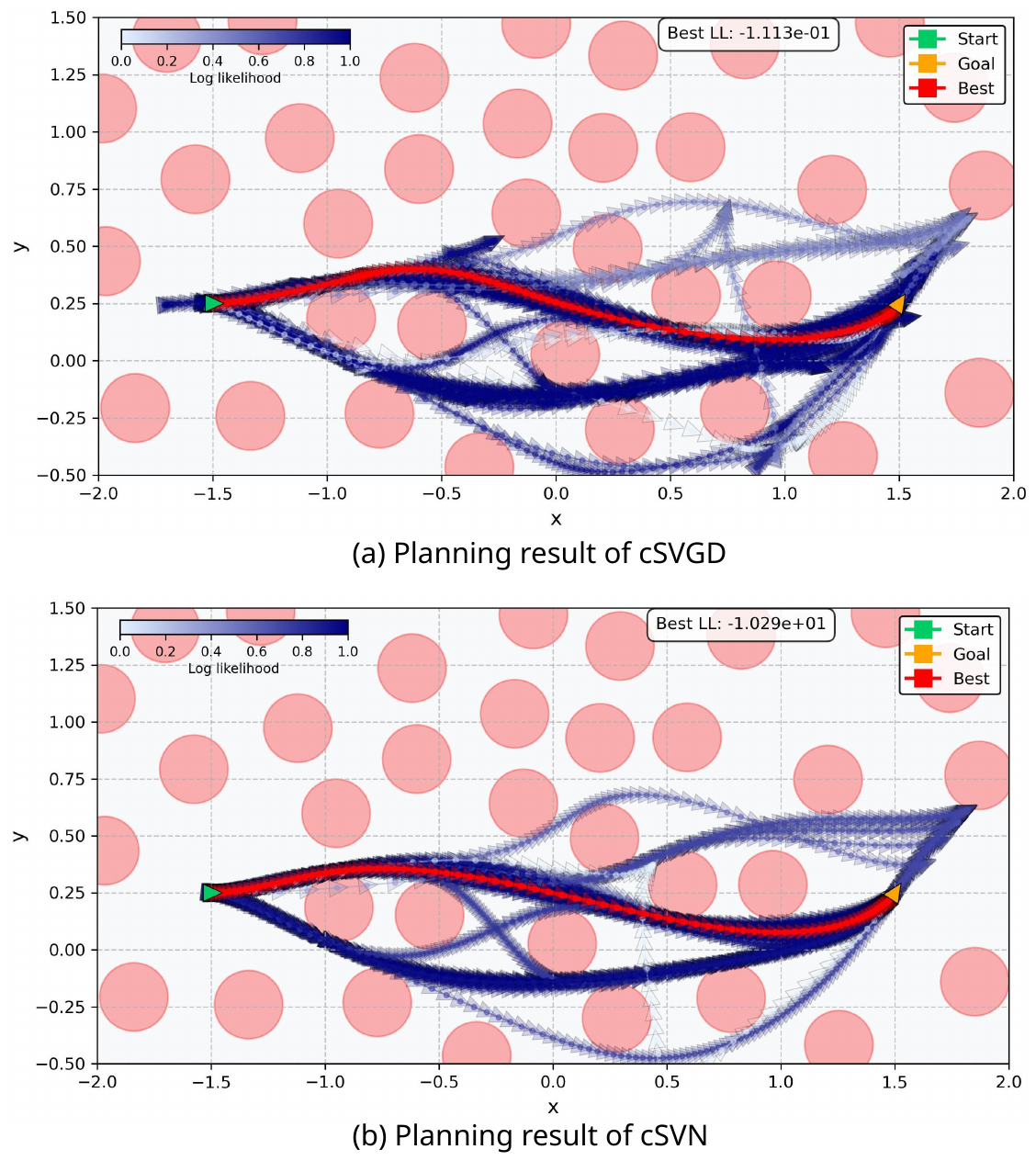}
    \caption{The unicycle problem involves maneuvering the unicycle from the starting pose, represented by the green triangle, to the target pose, represented by the yellow triangle. Pink circles indicate obstacles. Trajectories with high log-likelihood are shown in dark blue, while those with low log-likelihood appear in light blue. The best trajectory is highlighted in red. Both methods yield a similar distribution of particles that satisfy the non-holonomic constraints. However, cSVN identifies a trajectory with a higher log-likelihood than cSVGD.}
    \label{fig:final_planning_unicycle}
    \vspace{-14pt}
\end{figure}

Trajectory optimization for the unicycle navigation task is often challenging due to the need to satisfy non-holonomic constraints throughout the entire trajectory. An example of this problem setup is illustrated in Fig. \ref{fig:final_planning_unicycle}. The non-holonomic constraints $h(\xi_t)$ have the following form:
\begin{equation}
    \dot{y}_t \cos(\theta_t) - \dot{x}_t \sin(\theta_t ) = 0,
\end{equation}
where \( x_t \) and \( y_t \) represent the coordinates in 2D space at time $t$, and \( \theta_t \) denotes the direction of the unicycle.

\begin{figure}
    \centering
    \vspace{-10pt}
    \includegraphics[width=\linewidth]{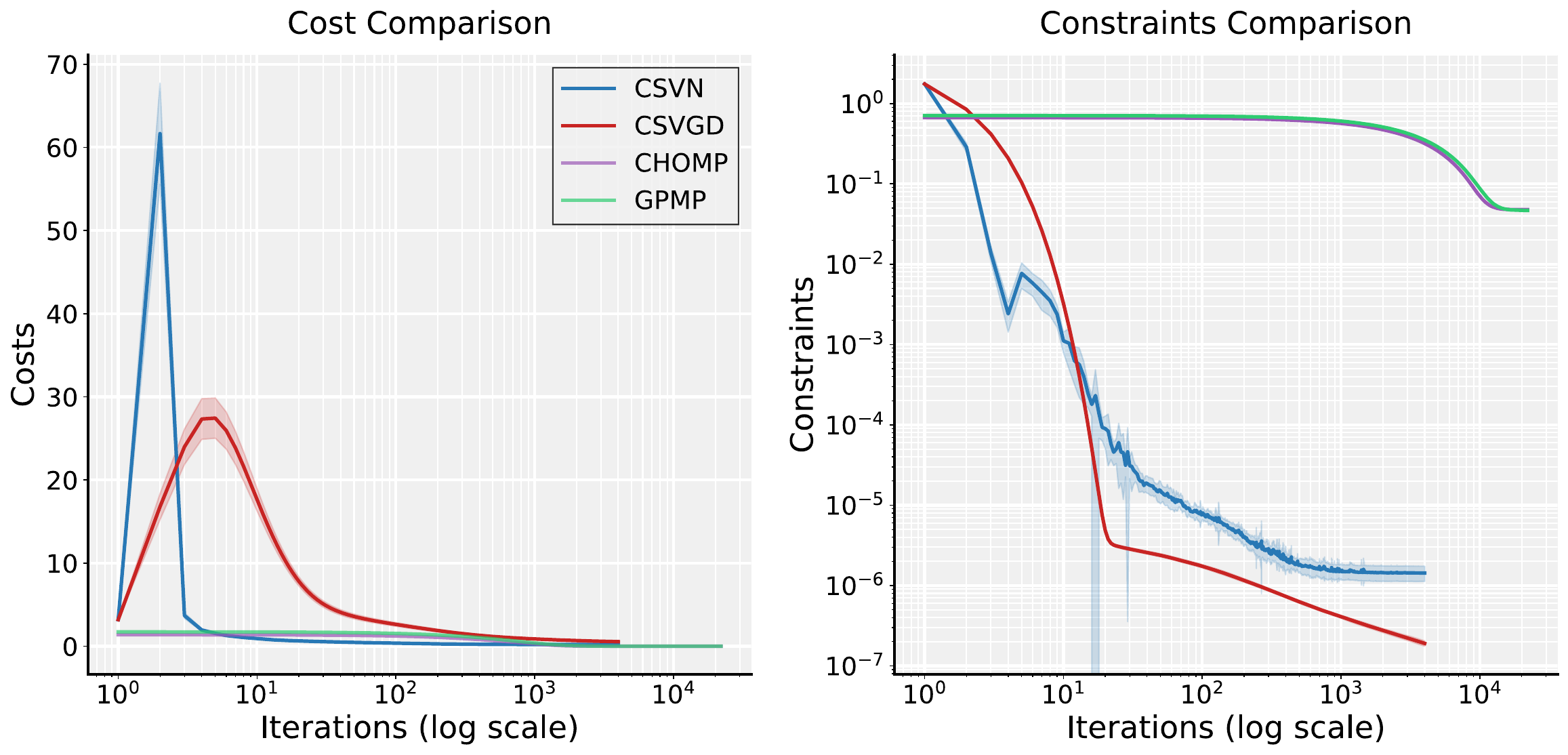}
    \caption{The left plot illustrates the convergence of the objective function, while the right plot depicts the convergence of the constraints. The constraint violation and iteration number are plotted on a logarithmic scale.}
    \label{fig:convergence_comparison}
    \vspace{-12pt}
\end{figure}

We compare cSVGD and cSVN, both designed to solve the proposed nonlinear GPMP inference problem, against CHOMP and GPMP, which incorporate a large squared penalty of \(100 h(\xi_t)^2\) for non-holonomic constraints. Each planner is executed ten times using the same set of random seeds with 50 particles. Since CHOMP and GPMP sample from different GP priors, their initial results are not directly aligned with those of cSVGD and cSVN. For cSVGD and cSVN, we set the update rates to 0.5 and 1.0, respectively. We observed that the large squared penalty violate the linear SDE prior requires CHOMP and GPMP to adopt a smaller update rate for stable convergence, leading us to set their learning rate to \(1e^{-4}\). These update rates were determined through test runs to empirically balance stability and efficiency. cSVGD and cSVN can operate with relatively large update rates because the nonlinear GPMP prior provides correct regularization, stabilizing the optimization process. In this comparison, the nonlinear GPMP regularization is scaled to \(1e^{-2}\). To account for differences in update rates, we run 4,000 steps for cSVN and cSVGD, while CHOMP and GPMP, which require smaller updates, are executed for 22,000 steps to achieve convergence. Figure \ref{fig:convergence_comparison} illustrates the convergence of the constraints and the objective function. The objective function scaling differs between CHOMP/GPMP and cSVGD/cSVN, as it was fine-tuned for optimal performance. We compute the mean and plot it alongside one standard deviation.

In the plot, the objective function values for cSVN and cSVGD initially increase over the first few iterations as the optimizer adjusts the initialization from the nonlinear GPMP prior— which does not satisfy the non-holonomic constraints—toward the constraint manifold. Notably, cSVN exhibits significantly faster convergence for the objective function compared to cSVGD. After just 80 iterations, cSVN achieves an objective value that cSVGD has yet to reach even after 4,000 iterations. This highlights that, on the non-holonomic constraint manifold, cSVN is at least 50 times more efficient than cSVGD in terms of problem query count. Despite this efficiency gap, both optimizers ultimately produce a similar final particle distribution, as shown in Fig. \ref{fig:final_planning_unicycle}. However, the results from CHOMP and GPMP indicate that, despite employing a large penalty function to enforce the non-holonomic constraints, these methods fail to fully adhere to the nonlinear constraints. Due to their inherent reliance on a linear SDE assumption, CHOMP and GPMP exhibit a substantial constraint violation of approximately \(5e^{-2}\) after convergence.

\subsection{Comparison with Standard Baselines on the Panda Robot Benchmark}
\begin{figure*}[t!]
    \centering
    \vspace{-5pt}
    \includegraphics[width=\textwidth]{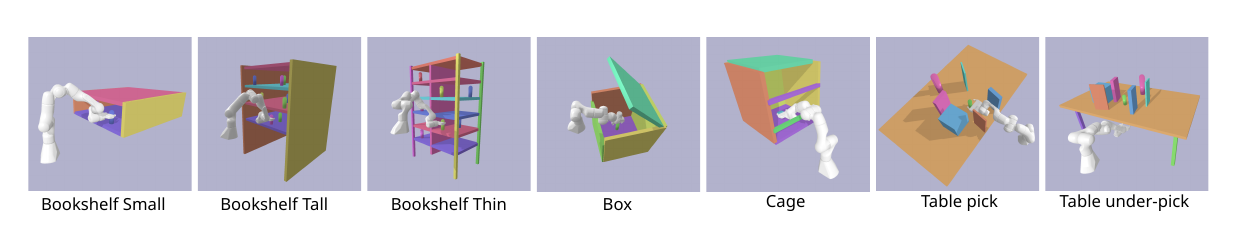}
    \caption{For the test benchmark, we show seven scenarios for robot grasping in different environments. These include three distinct bookshelf tasks, one box task, one cage task, and one table pick task. Additionally, we introduced a table under-pick task, which is identical to the table pick task except that the manipulator is initialized beneath the table. Most gradient-based local planners struggle to find a feasible path to the grasping configuration in this scenario.}
    \label{fig:combined_7_scenarios}
    \vspace{-5pt}
\end{figure*}

To demonstrate the ability of our planner to identify better local optima, we evaluate cSGPMP against two standard baselines GPMP and CHOMP using the MBM benchmark \cite{chamzas2021motionbenchmaker}, a grasping-focused dataset that requires a robot to transition from an initial configuration to a target grasping position while avoiding obstacles. The benchmark consists of seven scenarios, each containing 50 subproblems with varying initial and goal configurations shown in Fig. \ref{fig:combined_7_scenarios}. All planners are implemented in JAX, utilizing double-precision floating-point computations. Experiments are conducted on a PC equipped with an NVIDIA RTX 4080 GPU and an AMD Ryzen 9 7950X3D CPU.

For cSGPMP, we initialize $30$ particles from its GPMP trajectory prior and then perform variational inference. The 'best' trajectory is selected based on the lowest objective function value, evaluating the planners' ability to identify superior local optima. To ensure a fair comparison, we initialize the baseline GPMP and CHOMP with their respective trajectory priors, maintaining the same diversity in the initial particle distribution. Three levels of initialization diversity—small, medium, and large—are tested, with the best-performing setting used for comparison. The full results are provided in the Appendix. Success is determined based on collision avoidance, task constraint satisfaction, and adherence to joint limits. Constraint violations are quantified through target and goal position errors, while planners that lack hard joint constraint enforcement apply a large squared penalty for compliance. Additionally, all planners are subject to a shortest-path penalty. Although cSGPMP (full) strictly enforces constraints, we also include a variant, cSGPMP (eq), which treats joint limits as soft constraints using penalty costs instead of explicit inequalities with slack variables. Additionally, we found that cSGPMP effectively operates with the standard CHOMP cost function, which is characterized by a piecewise function with a gradient of constant magnitude. However, this cost function performs poorly for CHOMP and GPMP in scenarios involving narrow passages and small obstacles. These planners rely on a zero-acceleration prior, which smooths trajectories but fails to account for small cost variations. To enhance performance in such cases, we adopt a sharper cost function, \(\exp(-d)\), where \(d\) represents the distance to the obstacle surface.

We run all planners to convergence and evaluate the planning results using the following five criteria: (1) \textbf{Trajectory Length}: Sum of the norms of finite differences in the joint position trajectory. (2) \textbf{Smoothness}: Mean squared error (MSE) of finite differences in the velocity trajectory. (3) \textbf{Success Rate}: Ratio of successfully solved tasks (out of 50) per scenario.
(4) \textbf{Constraint Violations}: MSE used to indicate trajectory compliance with hard task constraints.
(5) \textbf{Computation Time}: Measured as the execution time after JIT compilation in JAX. The result is shown in Table \ref{tab:planner_comparison}.

\begin{table*}[htpb]
    \centering
    \renewcommand{\arraystretch}{0.1}
    \setlength{\tabcolsep}{10pt}
    \begin{tabular}{@{} lcccccr @{}}
        \toprule
        \textbf{Task} & \textbf{Planner} & \textbf{Time [s]} & \textbf{Success (\%)} & \textbf{Constraints} & \textbf{Length} & \textbf{Smoothness}\\
        \midrule
        \textbf{Bookshelf Small}      & CHOMP   & \textbf{0.839} ($\pm$0.002) & 78.00 & $1.26e^{-4}$ ($\pm$$1.97e^{-4}$) & \textbf{4.270} ($\pm$0.693) & \textbf{0.000} ($\pm$0.000) \\
                    & GPMP  & 0.942 ($\pm$0.002)  & 92.00 & $8.33e^{-5}$ ($\pm$3.49$e^{-5}$) & 4.937 ($\pm$0.596) & 0.028 ($\pm$0.004) \\
                    & cSGPMP (eq) & 1.086 ($\pm$0.021) & \textbf{100.00} & $6.27e^{-8}$ ($\pm$8.09$e^{-8}$) & 4.351 ($\pm$0.660) & 0.087 ($\pm$0.023) \\
                    & cSGPMP (full)  & 2.761 ($\pm$0.003)  & 98.00 & $\mathbf{2.17e^{-8}}$ ($\pm$7.02$e^{-8}$) & 4.351 ($\pm$0.657) & 0.083 ($\pm$0.022) \\
                    \midrule
        \textbf{Bookshelf Tall}  & CHOMP   & \textbf{1.401} ($\pm$0.004) & 88.00 & $9.68e^{-5}$ ($\pm$$3.85e^{-5}$) & \textbf{4.462} ($\pm$0.598) & \textbf{0.000} ($\pm$0.000) \\
                    & GPMP  & 1.490 ($\pm$0.003)  & 96.00 & $9.18e^{-5}$ ($\pm$3.64$e^{-5}$) & 5.077 ($\pm$0.559) & 0.029 ($\pm$0.004) \\
                    & cSGPMP (eq) & 1.523 ($\pm$0.040) & 96.00 & $4.10e^{-8}$ ($\pm$4.53$e^{-8}$) & 4.491 ($\pm$0.590) & 0.089 ($\pm$0.021) \\
                    & cSGPMP (full)  & 3.502 ($\pm$0.004)  & \textbf{98.00} & $\mathbf{6.49e^{-9}}$ ($\pm$2.09$e^{-8}$) & 4.495 ($\pm$0.582) & 0.090 ($\pm$0.023) \\
                    \midrule
        \textbf{Bookshelf Thin} & CHOMP   & 2.105 ($\pm$0.003) & 72.00 & $1.73e^{-4}$ ($\pm$$2.07e^{-4}$) & \textbf{4.305} ($\pm$0.529) & \textbf{0.000} ($\pm$0.000) \\
                    & GPMP  & 2.208 ($\pm$0.002)  & 88.00 & $9.19e^{-5}$ ($\pm$4.39$e^{-5}$) & 4.893 ($\pm$0.490) & 0.029 ($\pm$0.003) \\
                    & cSGPMP (eq) & \textbf{2.042} ($\pm$0.040) & \textbf{100.00} & $7.48e^{-8}$ ($\pm$6.71$e^{-8}$) & 4.328 ($\pm$0.486) & 0.086 ($\pm$0.016) \\
                    & cSGPMP (full)  & 4.391 ($\pm$0.003)  & \textbf{100.00} & $\mathbf{3.18e^{-8}}$ ($\pm$4.64$e^{-8}$) & 4.319 ($\pm$0.485) & 0.086 ($\pm$0.018) \\
        \midrule
        \textbf{Box} & CHOMP   & \textbf{1.029} ($\pm$0.001) & 54.00 & $7.26e^{-5}$ ($\pm$$2.34e^{-5}$) & 4.033 ($\pm$0.439) & \textbf{0.000} ($\pm$0.000) \\
                    & GPMP  & 1.117 ($\pm$0.001)  & 90.00 & $7.76e^{-5}$ ($\pm$3.76$e^{-5}$) & 4.618 ($\pm$0.393) & 0.028 ($\pm$0.004) \\
                    & cSGPMP (eq) & 1.208 ($\pm$0.001) & \textbf{100.00} & $6.56e^{-8}$ ($\pm$4.17$e^{-8}$) & 4.008 ($\pm$0.435) & 0.083 ($\pm$0.026) \\
                    & cSGPMP (full)  & 3.026 ($\pm$0.002)  & 98.00 & $\mathbf{2.75e^{-9}}$ ($\pm$4.46$e^{-10}$) & \textbf{3.965} ($\pm$0.396) & 0.075 ($\pm$0.022) \\
        \midrule
        \textbf{Cage}   & CHOMP   & \textbf{1.321} ($\pm$0.000) & 46.00 & $3.19e^{-4}$ ($\pm$$3.61e^{-4}$) & 5.663 ($\pm$0.611) & \textbf{0.000} ($\pm$0.000) \\
                    & GPMP  & 1.430 ($\pm$0.023)  & 38.00 & $9.85e^{-5}$ ($\pm$5.09$e^{-5}$) & 5.697 ($\pm$0.595) & 0.029 ($\pm$0.004) \\
                    & cSGPMP (eq) & 1.478 ($\pm$0.000) & \textbf{98.00} & $1.65e^{-7}$ ($\pm$1.20$e^{-7}$) & \textbf{5.051} ($\pm$0.480) & 0.117 ($\pm$0.020) \\
                    & cSGPMP (full)  & 3.521 ($\pm$0.001)  & 96.00 & $\mathbf{1.01e^{-7}}$ ($\pm$1.10$e^{-7}$) & 5.042 ($\pm$0.479) & 0.112 ($\pm$0.020) \\
        \midrule
        \textbf{Table pick} & CHOMP   & \textbf{1.614} ($\pm$0.002) & 77.55 & $1.09e^{-4}$ ($\pm$$1.23e^{-4}$) & \textbf{4.383} ($\pm$0.410) & \textbf{0.000} ($\pm$0.000) \\
                    & GPMP  & 1.719 ($\pm$0.002)  & 83.67 & $9.47e^{-5}$ ($\pm$2.67$e^{-5}$) & 5.035 ($\pm$0.368) & 0.030 ($\pm$0.004) \\
                    & cSGPMP (eq) & 1.627 ($\pm$0.026) & 95.92 & $3.06e^{-8}$ ($\pm$4.08$e^{-8}$) & 4.539 ($\pm$0.494) & 0.095 ($\pm$0.026) \\
                    & cSGPMP (full)  & 3.743 ($\pm$0.003)  & \textbf{100.00} & $\mathbf{4.88e^{-9}}$ ($\pm$8.20$e^{-9}$) & 4.560 ($\pm$0.489) & 0.092 ($\pm$0.026) \\
        \midrule
        \textbf{Table under-pick} & CHOMP   & \textbf{1.616} ($\pm$0.002) & 2.00 & $1.27e^{-4}$ ($\pm$$0.00e^{0}$) & 5.873 ($\pm$0.000) & \textbf{0.000} ($\pm$0.000) \\
                    & GPMP  & 1.722 ($\pm$0.002)  & 2.00 & $1.77e^{-4}$ ($\pm$0.00$e^{0}$) & 6.456 ($\pm$0.000) & 0.035 ($\pm$0.000) \\
                    & cSGPMP (eq) & 1.641 $(\pm$0.032) & 98.00 & $9.33e^{-8}$ ($\pm$$7.53e^{-8}$) & \textbf{4.676} ($\pm$2.044) & 0.144 ($\pm$0.075) \\
                    & cSGPMP (full)  & 3.746 ($\pm$0.003)  & \textbf{100.00} & $\mathbf{6.60e^{-9}}$ ($\pm$1.23$e^{-8}$) & 4.693 ($\pm$2.034) & 0.139 ($\pm$0.070) \\
        \midrule
        \textbf{Average success (\%)}& CHOMP& GPMP& cSGPMP (eq)& cSGPMP (full)&&\\
        \midrule
        \textbf{over all problems}&59.65&69.96&98.27&\textbf{98.57}&&\\
        \bottomrule
    \end{tabular}
    \caption{Comparison of cSGPMP, CHOMP, and GPMP on benchmark tasks. All values, except for success rate, represent mean values, with the corresponding standard deviations shown in parentheses. The smoothness indicator of CHOMP is smaller than $1e^{-3}$ thus shows $0.000$.}
    \label{tab:planner_comparison}
    \vspace{-5pt}
\end{table*}

In the \textbf{Cage} task—characterized by narrow passages and small obstacles—CHOMP and GPMP achieve low success rates ($46.00\%$ and $38.00\%$, respectively), whereas cSGPMP significantly outperforms them with success rates of 98.00\% (eq) and $96.00\%$ (full). This demonstrates cSGPMP's superior ability to navigate intricate environments with fine obstacle structures while maintaining low constraint violations and smooth trajectories.  
The \textbf{Table Under-Pick} task presents an even greater challenge, requiring planners to escape local minima while navigating among small obstacles. CHOMP and GPMP struggle, achieving only a $2.00\%$ success rate due to their strong dependence on local gradient-based optimization without leveraging mutual information between trajectories. We found that for GPMP and CHOMP, the quality of initialization heavily influences planning outcomes. 
To analyze the impact of different initializations on planning performance, we compare setups with small, medium, and large diversity, with full results provided in the Appendix. Results indicate that increasing initial diversity generally improves CHOMP and GPMP success rates in tasks requiring extensive exploration, as broader initialization helps escape poor local minima. However, cSGPMP exhibits minimal sensitivity to this variable, highlighting its robustness and reduced need for hyperparameter tuning. While GPMP and CHOMP show some performance improvements with better initialization, they still rarely succeed, whereas cSGPMP consistently achieves a success rate of about $98\%$.

cSGPMP (full) is slower due to its treatment of joint limits as general inequality constraints with slack variables. In contrast, cSGPMP (eq) achieves comparable speed to GPMP and CHOMP. In complex environments such as \textbf{Bookshelf Thin}, which contains numerous obstacles and problem query is expensive, cSGPMP (eq) converges faster due to requiring fewer iterations. Unlike most gradient-based methods that necessitate a small learning rate for convergence, cSGPMP employs an update rate of \(1\) with a constant Levenberg damping term to ensure the positive definiteness of the cSVN Hessian matrix. Constraint compliance analysis shows that cSGPMP achieves at least three orders of magnitude tighter adherence compared to baselines. For Bookshelf tasks, CHOMP and GPMP generate trajectories of comparable length performance to cSGPMP. However, in more complex scenarios—such as \textbf{Box}, \textbf{Cage}, and \textbf{Table Under-Pick}, where local minima play a significant role, CHOMP and GPMP struggle to find efficient paths, failing to compete with cSGPMP. In terms of smoothness, CHOMP and GPMP employ a strong zero-acceleration prior, yielding minimal acceleration variations. In contrast, cSGPMP, which lacks this prior assumption, allows greater joint acceleration variation, resulting in more dynamic trajectories.

Across all tasks, cSGPMP consistently outperforms CHOMP and GPMP in success rate. While CHOMP and GPMP achieve average success rates of $59.65\%$ and $69.96\%$, respectively, cSGPMP exceeds $98\%$, demonstrating superior robustness and generalizability. To further validate our method, we implemented it on a real robot, showcasing its ability to find multi-modal planning solutions even in narrow configurations. The problem presents two distinct trajectory modes, both of which our planner successfully identifies, as illustrated in Fig. \ref{fig:real_exp}. Additionally, the full trajectory bundle generated by our planner is provided in the Appendix.
\section{Limitations}
While our planner with cSVN effectively handles equality constraints, it requires the introduction of slack variables for optimization with inequality constraints. This significantly increases the dimensionality of the optimization problem, particularly in the context of trajectory optimization.

Furthermore, the use of a second-order method introduces additional numerical challenges. When solving the KKT system via the Schur complement, near-singular Jacobian matrices demand higher numerical precision for accurate matrix inversion. As a result, a double-precision float number is necessary in our implementation to ensure numerical stability.

Lastly, unlike CHOMP and GPMP, our approach does not incorporate a zero-acceleration or zero-jerk Gaussian process prior. Consequently, our planner tends to generate more dynamic trajectories, which, in some robotic applications, may be undesirable. These are topics that need to be addressed in future research.

\section{Conclusion} 
\label{sec:conclusion}
In this work, we introduced cSGPMP, a novel probabilistic inference framework that integrates a new GPMP trajectory prior and leverages second-order information to enable efficient variational inference for constrained trajectory optimization. Our framework significantly outperforms competitive baselines, including cSVGD, GPMP, and CHOMP, in both the nonholonomic constrained Unicycle problem and the Panda MBM benchmark.

In the Unicycle problem, cSGPMP with cSVN achieves at least $50$ times faster convergence than cSVGD on the constraint manifold, demonstrating its efficiency in optimization while strictly adhering to constraints through the use of second-order information. In the Panda MBM benchmark, our planner consistently outperforms CHOMP and GPMP with an overall $98.57\%$ success rate, showcasing its ability to identify better local optima, particularly in environments with narrow passages and intricate obstacle structures, where planners with random initialization struggle.

Overall, cSGPMP outperforms existing approaches by achieving higher success rates, improved objective function values, and greater computational efficiency for constrained motion planning in complex scenarios. This makes it a promising approach for challenging robotic planning problems. Future work will focus on extending this framework to dynamic environments and high-dimensional robotic systems, further enhancing its real-world applicability.

% \section*{Acknowledgments}

%% Use plainnat to work nicely with natbib. 

\bibliographystyle{unsrt}
\bibliography{references}

\clearpage
\section{Appendix}
\subsection{KKT with Slack Variables}
When handling nonlinear inequality constraints \( g(x) \leq 0 \), we can reformulate the original problem as an equality-constrained QP by introducing squared slack variables. By applying a Taylor expansion, this transformation guarantees equivalence between the original problem and the reformulated version presented below:
\begin{align}
    \argmin_{\delta_x, \delta_s} \quad & \mathcal{L}_1(\delta_x) + \mathcal{L}_2(\delta_x), \\
    \text{s.t.}\nonumber \quad \mathbb{E}_{y} &[\nabla h(y)^T \delta_x(y) + h(y)] = 0,\\
   \nonumber \quad \mathbb{E}_{y} &[\nabla g(y)^T \delta_x(y) + g(y) ] + \frac{1}{2} s^2 + \text{diag}(s) \delta_s = 0,
\end{align}
where $\delta_x$ represents the update direction of the original decision variable, and $\delta_s$ denotes the update direction of the slack variables. The corresponding KKT system is 
\label{eq:slack_KKT}
\begin{align}
    \setlength{\arraycolsep}{2pt} % Adjust column spacing
    \renewcommand{\arraystretch}{0.8} % Adjust row spacing
    \begin{bmatrix}
        \mathbf{H} & \mathbf{0} & \vline & \mu_p \nabla g & \mu_p \nabla h \\
        \mathbf{0} & \mathbf{0} & \vline & \mathbf{0} & \text{diag}(s) \\
        \hline
        \mu_p \nabla h^T & \mathbf{0} & \vline & \mathbf{0} & \mathbf{0} \\
        \mu_p \nabla g^T & \text{diag}(s) & \vline & \mathbf{0} & \mathbf{0}
    \end{bmatrix}
    \begin{bmatrix}
        \delta_x \\
        \delta_s \\
        \hline
        \boldsymbol{\lambda}_h \\
        \boldsymbol{\lambda}_g
    \end{bmatrix}
    =
    \begin{bmatrix}
        \boldsymbol{\phi} \\
        -\beta s \\
        \hline
        -h(y) \\
        -g(y) - \frac{1}{2}s^2
    \end{bmatrix}.
\end{align}
By solving the KKT conditions using a block matrix approach, we derive the update rules for both the particle and slack variables. To ensure strict positive definiteness, a damping term is added to the diagonal of the upper-left Hessian block. This damping term, relative to the diagonal zero matrix, acts as a squared penalty on the rate of change of the slack variables and, therefore, does not impact the final result.  

For implementation, we can leverage Cholesky decomposition. Specifically, we only need to compute the Cholesky decomposition for \( \mathbf{H} \) and the block diagonal matrix with the applied diagonal damping. The resulting Cholesky factor of \( \mathbf{H} \) can then be efficiently concatenated with the damping quotient to achieve the desired decomposition.

\subsection{Integration of HSGP features}
\label{subsec:hsgp_integration}
The key to deriving an analytically integrable kernel function is to represent it as the product of two variable-separated feature functions using the HSGP approach.

\begin{align}
    &\int_0^t \int_0^s k(\tau, \tau') d\tau d\tau',\\
    \nonumber = &\int_0^t \int_0^s \sum^m_{j=1} S(\sqrt{\lambda_j}) \varphi_j(\tau)\varphi_j(\tau')d\tau d\tau',\\
    \nonumber= &\sum_j S(\sqrt{\lambda_j}) [\int_0^t\varphi_j(\tau) d\tau] [\int_0^s \varphi_j(\tau') d\tau'].
\end{align}
and 
\begin{equation}
    \int_0^t k(\tau, s) d\tau = \sum_j S(\sqrt{\lambda_j}) [\int_0^t\varphi_j(\tau) d\tau] \varphi_j(s)
\end{equation}
For a Hilbert space reduced-rank kernel with symmetric zero-variance boundary conditions, the $\varphi_j$ feature and integral with respect to $\varphi_j$ are:
\begin{align}
    \varphi_j(x) &= \sqrt{\frac{1}{L}} \sin\left(\frac{\pi j (x + L)}{2L}\right) \\
    \int_0^t\varphi_j(\tau) d\tau &= \frac{1}{\sqrt{L}}[-\frac{2L}{\pi j} cos(\frac{\pi j(t+L)}{2L}) + \frac{2L}{\pi j} cos(\frac{\pi j}{2})],
\end{align}
where $L$ represents the radius of the active region for the approximated kernel function (e.g., $L=1$ for a kernel active only on $[-1,1]$). Beyond this region, the kernel produces zero variance, rendering it undefined in these areas. For more details, we refer to \cite{solin2020hilbert}.

\subsection{Derivation of the Numerical Realization of the cSGPMP}
\label{subsec:numerical_cSGMP}
In this section, we demonstrate that Algorithm \ref{alg:cSVN} provides a solution to the cSVN KKT system defined in Eq. \ref{eq:KKT_system}. For the update direction \(\delta\), we assume it takes the form of a function in the RKHS as \(\delta = \sum_{i=0}^d k(\cdot, x_i) \alpha_i\), which spans the RKHS \(\mathcal{H}^d\) with \(d\) basis functions. Each term in the KKT system \ref{eq:KKT_system} can be written using a set of \(N\) particles \(Q\). The empirical SVGD is
\begin{equation}
    \boldsymbol{\phi}(y) = \frac{1}{N} \sum^N_{i=1} \nabla_x \log p(x) k(y, x_i) + \nabla_x k(y,x_i).
\end{equation}
The evaluation of the inner product of Hessian and $\delta$ at $y$ is
\begin{equation}
    \langle \mathbf{H}, \delta \rangle_{\mathcal{H}}|_y = \sum_{i=1}^{d} \mathbf{H}(y,x_i) \alpha_i,
\end{equation}
where we use the fact that \(\mathbf{H}\) is a double-input operator and apply the reproducing property. The functional gradient spans the same dimension as the particle count. A key implication of the block-diagonal assumption is that the Hessian operator \(\mathbf{H}(x, y)\) returns \(0\) for \(x \neq y\). Therefore, the equation above simplifies to:
\begin{equation}
    \sum_{i=1}^{d} \mathbf{H}(y,x_i) \alpha_i = \mathbf{H}(y,y) \alpha_y,
\end{equation}
where empirical \(\mathbf{H}(y, y)\) is 
\begin{equation}
    \frac{1}{N}\sum^N_{i=1}[- \nabla^2 \log p(x_i) k(x_i,y)^2 +\nabla_x k(y,x_i) \nabla_x k(y,x_i)^T].
\end{equation}
Without considering the noise in $y$, we can write
\begin{equation}
    \mu_p \nabla h \, \boldsymbol{\lambda} = \nabla_y h(y) \, \boldsymbol{\lambda} k(y,y) = \nabla_y h(y) \, \boldsymbol{\lambda},
\end{equation}
where \(k(y, y) = 1\), a property generally satisfied by most stationary kernels. The primary condition is:
\begin{equation}
    \langle \mu_p \nabla h^T, \delta\rangle_{\mathcal{H}} = \sum^N_{i=1} k(y,x_i) \nabla h(y)^T \alpha_i \approx \nabla_y h(y) ^T \alpha_y,
\end{equation}
The term in the middle shows, if we didn't impose the constraints-to-particles independence, the constraints evaluated at $y$ will affect all particles through the connection of the kernel function. To preserve the particle-wise disentangled update direction introduced by the block-diagonal assumption, we assume that only the current particle is impacted by the constraints evaluated at its position. This leads to the above approximation. 

In this case, the KKT system can be reformulated to $N$ subproblems using the new set of parameters, \(\alpha_y\) and \(\boldsymbol{\lambda}\), within the framework of standard matrix algebra. The solution for \(\alpha_y\) can then be directly applied to update the corresponding particle.

\subsection{Point Mass Problem Solved by Three Planners}
\begin{figure}[htb]
    \centering
    \begin{subfigure}{\linewidth}
        \centering
        \includegraphics[width=\textwidth]{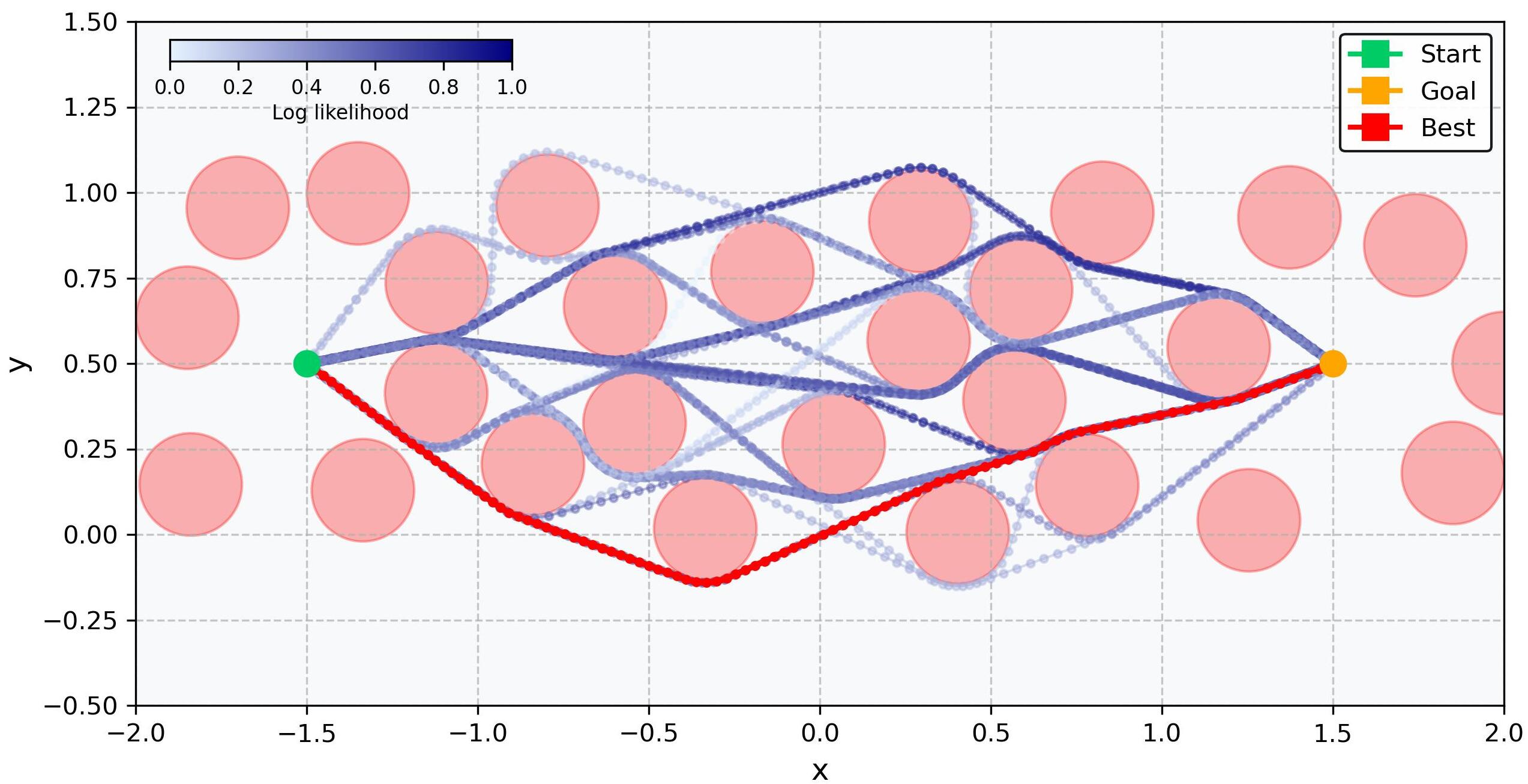}
        \caption{Result of CHOMP}
    \end{subfigure}
    \hfill
    \begin{subfigure}{\linewidth}
        \centering
        \includegraphics[width=\textwidth]{img/chomp_obs.jpg}
        \caption{Result of GPMP}
    \end{subfigure}
    \hfill
    \begin{subfigure}{\linewidth}
        \centering
        \includegraphics[width=\textwidth]{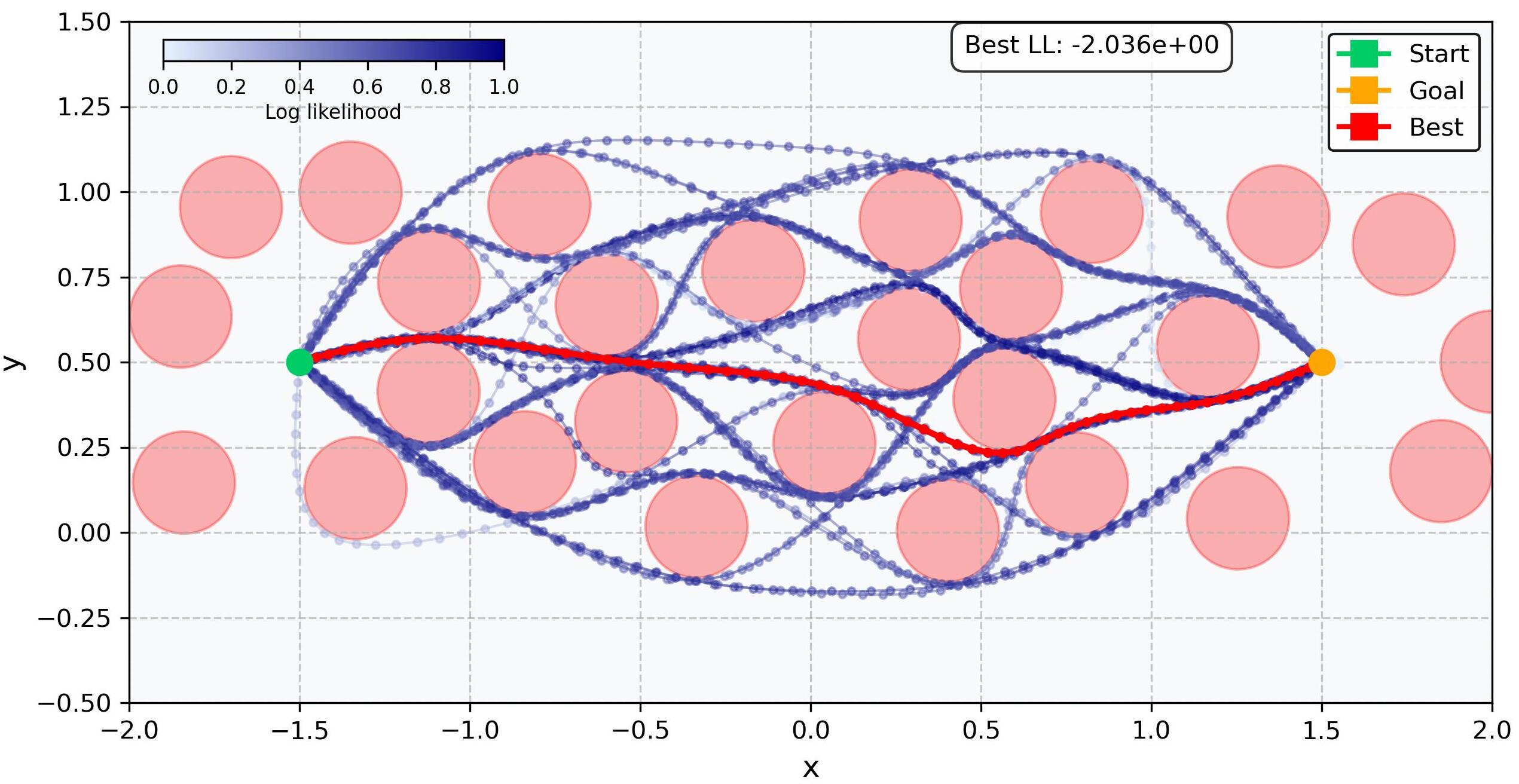}
        \caption{Result of cSGPMP}
    \end{subfigure}

    \caption{The results of three planners. The color of trajectories reflect the value of log-likelihood (cost function) and the best trajectory is choose with the smallest cost.}
    \label{fig:point_mass_three_planners}
\end{figure}

To gain deeper insight into the planning results of the three planners compared in Section \ref{sec:results}, we conducted an additional experiment where a point mass navigates through dense obstacles in a 2D space. The results, visualized with 50 trajectories in Fig. \ref{fig:point_mass_three_planners}, highlight the distinct characteristics of each planner. Notably, CHOMP produces the smoothest trajectories, but this smoothness comes at the cost of limited state-space exploration. GPMP exhibits a similar limitation; however, due to its regularization by a linear GP prior, its results are strongly influenced by this prior. In contrast, cSGPMP, leveraging the impact of SVN, achieves better trajectory mode coverage, effectively balancing smoothness and exploration.

\subsection{Comparison with sampling from different GP variance}
The results presented in \ref{tab:planner_sigma_comparison_csgpmp} demonstrate the performance of the planner across different values of sigma value for initialization of the GP prior, indicating different intialization of diversity (small, medium, large) with the success rates reflecting the trade-off between trajectory quality and constraint satisfaction. Specifically, for sigma = 5, the overall success rate is 98.57\%, while for sigma = 0.5, the success rate slightly decreases to 98.00\%. For sigma = 2.5, the success rate further drops to 97.43\%. These findings highlight the influence of sigma on trajectory planning: smaller sigma values result in smoother and shorter trajectories, thereby improving trajectory quality. Conversely, larger sigma values enhance the satisfaction of constraints, albeit at the cost of reduced trajectory quality. However, we observed that the choice of these hyperparameters had a minor impact on overall performance. We also identified different sigma value for CHOMP and GPMP w.r.t. small, medium and large initialization diversity, shown in Table \ref{tab:planner_sigma_comparison_chomp} and Table \ref{tab:planner_sigma_comparison_gpmp}.

\begin{table*}[htpb]
    \centering
    \renewcommand{\arraystretch}{0.5} % Adjust row spacing
    \setlength{\tabcolsep}{5pt} % column spacing
    \begin{tabular}{@{} lcccccr @{}}
        \toprule
        \textbf{Tasks} & \textbf{Planner} & \textbf{Time [s]} & \textbf{Success (\%)} & \textbf{Constraints} & \textbf{Length} & \textbf{Smoothness}\\
        \midrule
        \textbf{Bookshelf Small}      & cSGPMP (sigma 0.5)   & 2.839 (0.061) & \textbf{100.00} & $1.33e^{-7}$ ($1.21e^{-7}$) & 4.224 (0.680) & 0.071 (0.023) \\
                    & cSGPMP (sigma 2.5)  & 2.768 (0.003)  & 98.00 & $\mathbf{1.87e^{-8}}$ ($4.91e^{-8}$) & \textbf{4.220} (0.691) & \textbf{0.069} (0.024) \\
                    & cSGPMP (sigma 5.0)  & \textbf{2.761} (0.003)  & 98.00 & $2.17e^{-8}$ ($7.02e^{-8}$) & 4.351 (0.657) & 0.083 (0.022) \\
                    \midrule
        \textbf{Bookshelf Tall}  & cSGPMP (sigma 0.5)   & 3.641 (0.096) & 96.00 & $1.17e^{-7}$ ($4.13e^{-8}$) & 4.351 (0.606) & \textbf{0.072} (0.019) \\
                                & cSGPMP (sigma 2.5)  & 3.587 (0.065)  & 96.00 & $7.50e^{-9}$ ($2.95e^{-9}$) & \textbf{4.343} (0.606) & 0.072 (0.020) \\
                                & cSGPMP (sigma 5.0)  & \textbf{3.502} (0.004)  & \textbf{98.00} & $\mathbf{6.49e^{-9}}$ ($2.09e^{-8}$) & 4.495 (0.582) & 0.090 (0.023) \\
                    \midrule
        \textbf{Bookshelf Thin} & cSGPMP (sigma 0.5)   & 4.535 (0.095) & \textbf{100.00} & $1.59e^{-7}$ ($9.04e^{-8}$) & 4.182 (0.505) & 0.069 (0.017) \\
                                & cSGPMP (sigma 2.5)  & 4.489 (0.059)  & \textbf{100.00} & $3.83e^{-8}$ ($5.09e^{-8}$) & \textbf{4.176} (0.508) & \textbf{0.067} (0.017) \\
                                & cSGPMP (sigma 5.0)  & \textbf{4.391} (0.003)  & \textbf{100.00} & $\mathbf{3.18e^{-8}}$ ($4.64e^{-8}$) & 4.319 (0.485) & 0.086 (0.018) \\
        \midrule
        \textbf{Box} & cSGPMP (sigma 0.5)   & 3.152 (0.047) & 96.00 & $8.30e^{-8}$ ($1.68e^{-8}$) & \textbf{3.830} (0.401) & \textbf{0.058} (0.013) \\
                    & cSGPMP (sigma 2.5)  & 3.052 (0.038)  & 96.00 & $5.84e^{-9}$ ($1.07e^{-9}$) & \textbf{3.830} (0.401) & \textbf{0.058} (0.013) \\
                    & cSGPMP (sigma 5.0) & \textbf{3.026} (0.002)  & \textbf{98.00} & $\mathbf{2.75e^{-9}}$ ($4.46e^{-10}$) & 3.965 (0.396) & 0.075 (0.022) \\
        \midrule
        \textbf{Cage}   & cSGPMP (sigma 0.5)   & 3.679 (0.051) & 96.00 & $3.12e^{-7}$ ($2.03e^{-7}$) & \textbf{4.969} (0.495) & \textbf{0.100} (0.018) \\
                        & cSGPMP (sigma 2.5)  & 3.563 (0.031)  & \textbf{98.00} & $1.17e^{-7}$ ($1.24e^{-7}$) & 4.980 (0.484) & 0.103 (0.017) \\
                        & cSGPMP (sigma 5.0)  & \textbf{3.521} (0.001)  & 96.00 & $\mathbf{1.01e^{-7}}$ ($1.10e^{-7}$) & 5.042 (0.479) & 0.112 (0.020) \\
        \midrule
        \textbf{Table pick} & cSGPMP (sigma 0.5)   & 3.946 (0.095) & \textbf{100.00} & $1.17e^{-7}$ ($3.13e^{-8}$) & 4.437 (0.507) & \textbf{0.078} (0.025) \\
                            & cSGPMP (sigma 2.5)  & 3.794 (0.028)  & \textbf{100.00} & $8.88e^{-9}$ ($7.41e^{-9}$) & \textbf{4.432} (0.510) & 0.078 (0.027) \\
                            & cSGPMP (sigma 5.0)  & \textbf{3.743} (0.003)  & \textbf{100.00} & $\mathbf{4.88e^{-9}}$ ($8.20e^{-9}$) & 4.560 (0.489) & 0.092 (0.026) \\
        \midrule
        \textbf{Table under-pick} & cSGPMP (sigma 0.5)   & 3.886 (0.086) & 98.00 & $1.41e^{-7}$ ($1.02e^{-7}$) & \textbf{4.588} (2.067) & \textbf{0.122} (0.074) \\
                                & cSGPMP (sigma 2.5)  & 3.838 (0.074)  & 94.00 & $1.20e^{-8}$ ($1.35e^{-8}$) & 4.621 (2.028) & 0.126 (0.067) \\
                                & cSGPMP (sigma 5.0)  & \textbf{3.746} (0.003)  & \textbf{100.00} & $\mathbf{6.60e^{-9}}$ ($1.23e^{-8}$) & 4.693 (2.034) & 0.139 (0.070) \\
        \midrule
        \textbf{Average success (\%)}& cSGPMP (sigma 0.5)& cSGPMP (sigma 2.5)& cSGPMP (sigma 5.0)& &&\\
        \midrule
        \textbf{over all problems}&98\%&97.43\%&\textbf{98.57}\%&&&\\
        \bottomrule
    \end{tabular}
    \caption{Comparison of different initialization diversity for cSGPMP on benchmark tasks. Small: sigma=0.5, Medium: sigma=2.5, Large: sigma=5.0. All values, except for success rate, represent mean values, with the corresponding standard deviations shown in parentheses.}
    \label{tab:planner_sigma_comparison_csgpmp}
\end{table*}

\begin{table*}[htpb]
    \centering
    \renewcommand{\arraystretch}{0.5} % Adjust row spacing
    \setlength{\tabcolsep}{5pt} % column spacing
    \begin{tabular}{@{} lcccccr @{}}
        \toprule
        \textbf{Tasks} & \textbf{Planner} & \textbf{Time [s]} & \textbf{Success (\%)} & \textbf{Constraints} & \textbf{Length} & \textbf{Smoothness}\\
        \midrule
        \textbf{Bookshelf Small}      & GPMP (sigma 1.0)   & \textbf{0.943} (0.002) & 66.00 & $\mathbf{7.49e^{-5}}$ ($2.98e^{-5}$) & \textbf{4.089} (0.716) & \textbf{0.001} (0.001) \\
                    & GPMP (sigma 5.0)  & \textbf{0.943} (0.002)  & 88.00 & $7.88e^{-5}$ ($2.67e^{-5}$) & 4.479 (0.645) & 0.015 (0.002) \\
                    & GPMP (sigma 10.0)  & 0.944 (0.002)  & \textbf{90.00} & $9.32e^{-5}$ ($3.70e^{-5}$) & 5.317 (0.584) & 0.040 (0.005) \\
                    \midrule
        \textbf{Bookshelf Tall}  & GPMP (sigma 1.0)   & 1.494 (0.004) & 84.00 & $8.72e^{-5}$ ($2.34e^{-5}$) & \textbf{4.378} (0.621) & \textbf{0.001} (0.001) \\
                                & GPMP (sigma 5.0)  & \textbf{1.493} (0.003)  & 90.00 & $\mathbf{8.48e^{-5}}$ ($2.91e^{-5}$) & 4.633 (0.599) & 0.015 (0.002) \\
                                & GPMP (sigma 10.0)  & \textbf{1.493} (0.003)  & \textbf{94.00} & $9.70e^{-5}$ ($4.84e^{-5}$) & 5.468 (0.563) & 0.040 (0.006) \\
                    \midrule
        \textbf{Bookshelf Thin} & GPMP (sigma 1.0)   & 2.216 (0.003) & 50.00 & $\mathbf{8.06e^{-5}}$ ($2.91e^{-5}$) & \textbf{4.083} (0.542) & \textbf{0.002} (0.001) \\
                                & GPMP (sigma 5.0)  & 2.216 (0.003)  & 82.00 & $8.76e^{-5}$ ($3.43e^{-5}$) & 4.426 (0.487) & 0.015 (0.002) \\
                                & GPMP (sigma 10.0)  & \textbf{2.215} (0.002)  & \textbf{92.00} & $9.91e^{-5}$ ($5.42e^{-5}$) & 5.294 (0.497) & 0.041 (0.005) \\
        \midrule
        \textbf{Box} & GPMP (sigma 1.0)   & 1.122 (0.001) & 48.00 & $\mathbf{6.79e^{-5}}$ ($1.88e^{-5}$) & \textbf{3.862} (0.411) & \textbf{0.003} (0.002) \\
                    & GPMP (sigma 5.0)  & \textbf{1.121} (0.001)  & \textbf{86.00} & $8.34e^{-5}$ ($3.01e^{-5}$) & 4.190 (0.390) & 0.015 (0.003) \\
                    & GPMP (sigma 10.0) & \textbf{1.121} (0.001)  & 82.00 & $7.14e^{-5}$ ($3.35e^{-5}$) & 5.111 (0.407) & 0.040 (0.005) \\
        \midrule
        \textbf{Cage}   & GPMP (sigma 1.0)   & 1.424 (0.001) & 32.00 & $1.14e^{-4}$ ($2.79e^{-5}$) & \textbf{5.165} (0.552) & \textbf{0.003} (0.001) \\
                        & GPMP (sigma 5.0)  & 1.423 (0.001)  & \textbf{58.00} & $1.25e^{-4}$ ($4.75e^{-5}$) & 5.332 (0.555) & 0.015 (0.002) \\
                        & GPMP (sigma 10.0)  & \textbf{1.422} (0.000)  & 20.00 & $\mathbf{9.32e^{-5}}$ ($6.31e^{-5}$) & 5.795 (0.549) & 0.039 (0.005) \\
        \midrule
        \textbf{Table pick} & GPMP (sigma 1.0)   & 1.719 (0.002) & 55.10 & $\mathbf{7.99e^{-5}}$ ($2.03e^{-5}$) & \textbf{4.219} (0.316) & \textbf{0.001} (0.000) \\
                            & GPMP (sigma 5.0)  & \textbf{1.718} (0.002)  & 81.63 & $8.38e^{-5}$ ($2.38e^{-5}$) & 4.630 (0.382) & 0.015 (0.002) \\
                            & GPMP (sigma 10.0)  & \textbf{1.718} (0.002)  & \textbf{87.76} & $1.02e^{-4}$ ($2.96e^{-5}$) & 5.496 (0.444) & 0.043 (0.005) \\
        \midrule
        \textbf{Table under-pick} & GPMP (sigma 1.0)   & 1.722 (0.002) & 0.00 & - & - & - \\
                                & GPMP (sigma 5.0)  & 1.722 (0.002)  & 2.00 & $1.45e^{-4}$ ($0.00e^{0}$) & 6.094 (0.000) & \textbf{0.017} (0.000) \\
                                & GPMP (sigma 10.0)  & \textbf{1.721} (0.002)  & \textbf{4.00} & $\mathbf{1.03e^{-4}}$ ($8.98e^{-5}$) & \textbf{4.886} (2.672) & 0.041 (0.007) \\
        \midrule
        \textbf{Average success (\%)}& GPMP (sigma 1.0)& GPMP (sigma 5.0)& GPMP (sigma 10.0)& &&\\
        \midrule
        \textbf{over all problems}&47.87\%&\textbf{69.66\%}&67.11\%&&&\\
        \bottomrule
    \end{tabular}
    \caption{Comparison of different initialization diversity for GPMP on benchmark tasks. Small: sigma=1.0, Medium: sigma=5.0, Large: sigma=10.0. All values, except for success rate, represent mean values, with the corresponding standard deviations shown in parentheses.}
    \label{tab:planner_sigma_comparison_gpmp}
\end{table*}

\begin{table*}[htpb]
    \centering
    \renewcommand{\arraystretch}{0.5} % Adjust row spacing
    \setlength{\tabcolsep}{5pt} % column spacing
    \begin{tabular}{@{} lcccccr @{}}
        \toprule
        \textbf{Tasks} & \textbf{Planner} & \textbf{Time [s]} & \textbf{Success (\%)} & \textbf{Constraints} & \textbf{Length} & \textbf{Smoothness}\\
        \midrule
        \textbf{Bookshelf Small}      & CHOMP (sigma 0.05)   & \textbf{0.839} (0.002) & 78.00 & \textbf{$\mathbf{1.26e^{-4}}$} ($1.97e^{-4}$) & \textbf{4.270} (0.693) & 0.000 (0.000) \\
                    & CHOMP (sigma 0.1)  & 0.839 (0.002)  & \textbf{78.00} & $1.84e^{-4}$ ($2.56e^{-4}$) & 4.500 (0.703) & 0.000 (0.000) \\
                    & CHOMP (sigma 0.5)  & 0.845 (0.001)  & 56.00 & $4.87e^{-4}$ ($4.56e^{-4}$) & 5.623 (1.187) & 0.000 (0.000) \\
                    \midrule
        \textbf{Bookshelf Tall}  & CHOMP (sigma 0.05)   & 1.401 (0.004) & 88.00 & \textbf{$\mathbf{9.68e^{-5}}$} ($3.85e^{-5}$) & \textbf{4.462} (0.598) & 0.000 (0.000) \\
                                & CHOMP (sigma 0.1)  & \textbf{1.394} (0.003)  & \textbf{92.00} & $1.14e^{-4}$ ($4.91e^{-5}$) & 4.560 (0.569) & 0.000 (0.000) \\
                                & CHOMP (sigma 0.5)  & 1.395 (0.001)  & 58.00 & $2.15e^{-4}$ ($7.96e^{-5}$) & 5.420 (0.678) & 0.000 (0.000) \\
                    \midrule
        \textbf{Bookshelf Thin} & CHOMP (sigma 0.05)   & \textbf{2.105} (0.003) & 72.00 & \textbf{$\mathbf{1.73e^{-4}}$} ($2.07e^{-4}$) & \textbf{4.305} (0.529) & 0.000 (0.000) \\
                                & CHOMP (sigma 0.1)  & 2.108 (0.003)  & \textbf{88.00} & $1.81e^{-4}$ ($1.52e^{-4}$) & 4.474 (0.527) & 0.000 (0.000) \\
                                & CHOMP (sigma 0.5)  & 2.108 (0.001)  & 26.00 & $3.99e^{-4}$ ($3.46e^{-4}$) & 5.572 (0.927) & 0.000 (0.000) \\
        \midrule
        \textbf{Box} & CHOMP (sigma 0.05)   & 1.029 (0.001) & \textbf{54.00} & \textbf{$\mathbf{7.26e^{-5}}$} ($2.34e^{-5}$) & \textbf{4.033} (0.439) & 0.000 (0.000) \\
                    & CHOMP (sigma 0.1)  & 1.024 (0.001)  & 50.00 & $9.11e^{-5}$ ($3.34e^{-5}$) & 4.162 (0.327) & 0.000 (0.000) \\
                    & CHOMP (sigma 0.5) & \textbf{1.018} (0.003)  & 0.00 & - & - & - \\
        \midrule
        \textbf{Cage}   & CHOMP (sigma 0.05)   & 1.321 (0.000) & \textbf{46.00} & $3.19e^{-4}$ ($3.61e^{-4}$) & 5.663 (0.611) & 0.000 (0.000) \\
                        & CHOMP (sigma 0.1)  & 1.319 (0.000)  & 4.00 & \textbf{$\mathbf{1.38e^{-4}}$} ($8.80e^{-5}$) & \textbf{5.168} (0.307) & 0.000 (0.000) \\
                        & CHOMP (sigma 0.5)  & \textbf{1.318} (0.000)  & 0.00 & - & - & - \\
        \midrule
        \textbf{Table pick} & CHOMP (sigma 0.05)   & \textbf{1.614} (0.002) & 77.55 & \textbf{$\mathbf{1.09e^{-4}}$} ($1.23e^{-4}$) & \textbf{4.383} (0.410) & 0.000 (0.000) \\
                            & CHOMP (sigma 0.1)  & 1.618 (0.002)  & \textbf{79.59} & $1.26e^{-4}$ ($1.28e^{-4}$) & 4.530 (0.428) & 0.000 (0.000) \\
                            & CHOMP (sigma 0.5)  & 1.617 (0.002)  & 20.41 & $3.53e^{-4}$ ($2.29e^{-4}$) & 5.550 (1.075) & 0.000 (0.000) \\
        \midrule
        \textbf{Table under-pick} & CHOMP (sigma 0.05)   & \textbf{1.616} (0.002) & 2.00 & \textbf{$\mathbf{1.27e^{-4}}$} ($0.00e^{0}$) & \textbf{5.873} (0.000) & 0.000 (0.000) \\
                                & CHOMP (sigma 0.1)  & 1.617 (0.002)  & 4.00 & $2.36e^{-4}$ ($1.08e^{-4}$) & 6.868 (1.254) & 0.000 (0.000) \\
                                & CHOMP (sigma 0.5)  & 1.619 (0.002)  & \textbf{6.00} & $4.75e^{-4}$ ($2.73e^{-4}$) & 7.607 (1.281) & 0.000 (0.000) \\
        \midrule
        \textbf{Average success (\%)}& CHOMP (sigma 0.05)& CHOMP (sigma 0.1)& CHOMP (sigma 0.5)& &&\\
        \midrule
        \textbf{over all problems}& \textbf{59.65\%} & 56.51\% & 23.77\% &&&\\
        \bottomrule
    \end{tabular}
    \caption{Comparison of different initialization diversity for CHOMP on benchmark tasks. Small: sigma=0.05, Medium: sigma=0.1, Large: sigma=0.5. All values, except for success rate, represent mean values, with the corresponding standard deviations shown in parentheses.}
    \label{tab:planner_sigma_comparison_chomp}
\end{table*}

\subsection{Trajectory Bundle for the problem in Fig. \ref{fig:real_exp}}
\begin{figure}
    \centering
    \includegraphics[width=\linewidth]{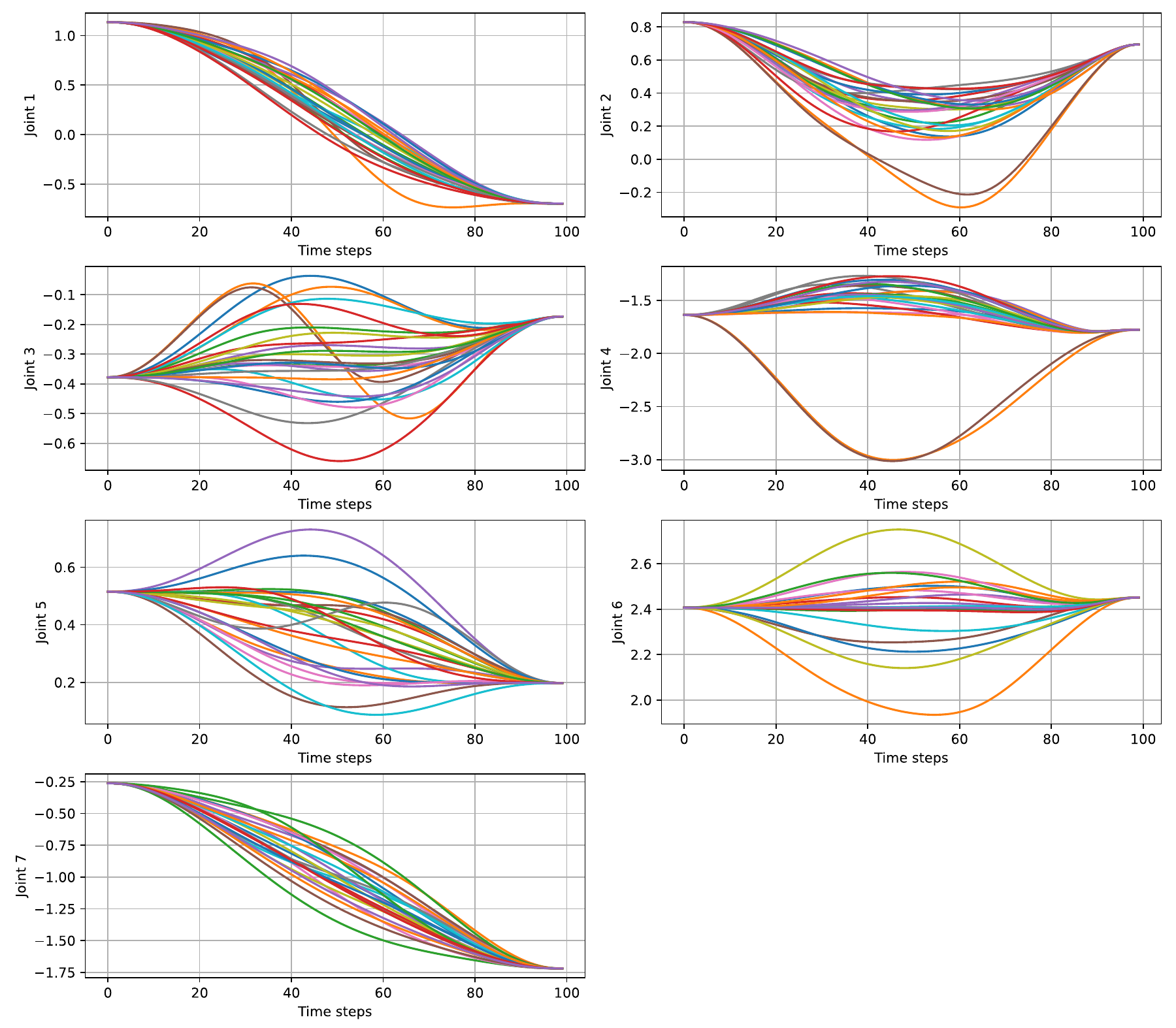}
    \caption{The planning result for the pick-in-box task with obstacle avoidance shows two distinct trajectories that navigate around the box rather than moving over it. The trajectory shows a shape of a distribution with mutiple modes.}
    \label{fig:trajectory_bundle}
\end{figure}

To demonstrate that our planner, driven by the Stein kernel force, identifies the trajectory as a distribution rather than a point estimate, we present the trajectory bundle from the real-life experiment, as shown in Fig. \ref{fig:trajectory_bundle}.
\end{document}